\pdfoutput=1
\documentclass[10pt,twocolumn,letterpaper]{article}

\usepackage[pagenumbers]{cvpr} 
\usepackage[accsupp]{axessibility}
\usepackage{lipsum}
\usepackage{listings}
\usepackage{subcaption}
\usepackage{ragged2e}
\usepackage{array}
\usepackage{times}
\usepackage{epsfig}
\usepackage{graphicx}
\usepackage{float}
\usepackage{marvosym}
\usepackage{wrapfig}
\usepackage{amsmath,amssymb,amsthm}
\usepackage{algorithm,algorithmicx,algpseudocode}
\usepackage{tcolorbox}
\usepackage{bm,xspace}
\usepackage{comment}
\usepackage{multirow}
\usepackage{balance}
\usepackage{url}
\usepackage{booktabs}
\usepackage{etoolbox,siunitx}
\usepackage{calc}
\usepackage{pifont,hologo}
\usepackage{color}
\usepackage{adjustbox}
\usepackage{amsmath}
\usepackage{enumitem}
\usepackage[normalem]{ulem}  %
\usepackage{contour}
\usepackage{colortbl} 
\usepackage{soul}%
\usepackage{tocloft} %
\usepackage{etoc} %
\PassOptionsToPackage{dvipsnames}{xcolor}
\usepackage[pagebackref,breaklinks,colorlinks,linkcolor=link,citecolor=cite]{hyperref}
\usepackage{natbib}
\PassOptionsToPackage{square,numbers,comma,sort}{natbib}

\newcommand{\ours}{Lyra}

\definecolor{mygreen}{HTML}{d0f4de}
\definecolor{myblue}{HTML}{50ACE9}
\newlength\savewidth\newcommand\shline{\noalign{\global\savewidth\arrayrulewidth\global\arrayrulewidth 1pt}\hline\noalign{\global\arrayrulewidth\savewidth}}
\newcommand{\cgaphl}[2]{\fontsize{7pt}{1em}\selectfont{\textcolor{red}{($\textbf{#1}$\textbf{#2})}}}
\newcommand{\cgaphll}[2]{\fontsize{7pt}{1em}\selectfont{\textcolor{Highlight}{($\textbf{#1}$\textbf{#2})}}}
\definecolor{Highlight}{HTML}{39b54a}

\definecolor{blue}{HTML}{004bb3}
\definecolor{red}{HTML}{cc1100}
\definecolor{orange}{HTML}{cc7700}
\definecolor{gray}{HTML}{efefef}
\definecolor{darkgreen}{HTML}{228B22}
\definecolor{darkgray}{HTML}{757575}

\definecolor{cite}{HTML}{3270b5}
\definecolor{link}{HTML}{b53532}
\definecolor{link}{HTML}{cc1100}
\definecolor{scratch}{HTML}{001219}
\definecolor{pretrain}{HTML}{0A9396}

\contourlength{0.8pt}

\renewcommand{\eqref}[1]{Eq.~\ref{#1}}

\newcolumntype{x}[1]{>{\centering\arraybackslash}p{#1}}
\newcolumntype{y}[1]{>{\raggedright\arraybackslash}p{#1}}
\newcolumntype{z}[1]{>{\raggedleft\arraybackslash}p{#1}}
\newcommand{\tablestyle}[2]{\setlength{\tabcolsep}{#1}\renewcommand{\arraystretch}{#2}\centering\footnotesize}

\DeclareMathSymbol{@}{\mathord}{letters}{"3B}

\newcommand{\cmark}{\ding{51}}%
\newcommand{\xmark}{\ding{55}}%

\makeatletter
\DeclareRobustCommand\onedot{\futurelet\@let@token\@onedot}
\def\@onedot{\ifx\@let@token.\else.\null\fi\xspace}
 
\def\ie{\emph{i.e}\onedot}

\newcommand*{\Rom}[1]{\expandafter\@slowromancap\romannumeral #1@}
\newcommand*{\rom}[1]{\expandafter\romannumeral #1}

\def\1{\bm{1}}

\DeclareMathAlphabet{\mathsfit}{\encodingdefault}{\sfdefault}{m}{sl}
\SetMathAlphabet{\mathsfit}{bold}{\encodingdefault}{\sfdefault}{bx}{n}

\let\originalleft\left
\let\originalright\right
\renewcommand{\left}{\mathopen{}\mathclose\bgroup\originalleft}
\renewcommand{\right}{\aftergroup\egroup\originalright}

\def\paperID{****} 
\def\confName{CVPR}
\def\confYear{2025}

\begin{document}
	
	\title{\includegraphics[width=0.45cm]{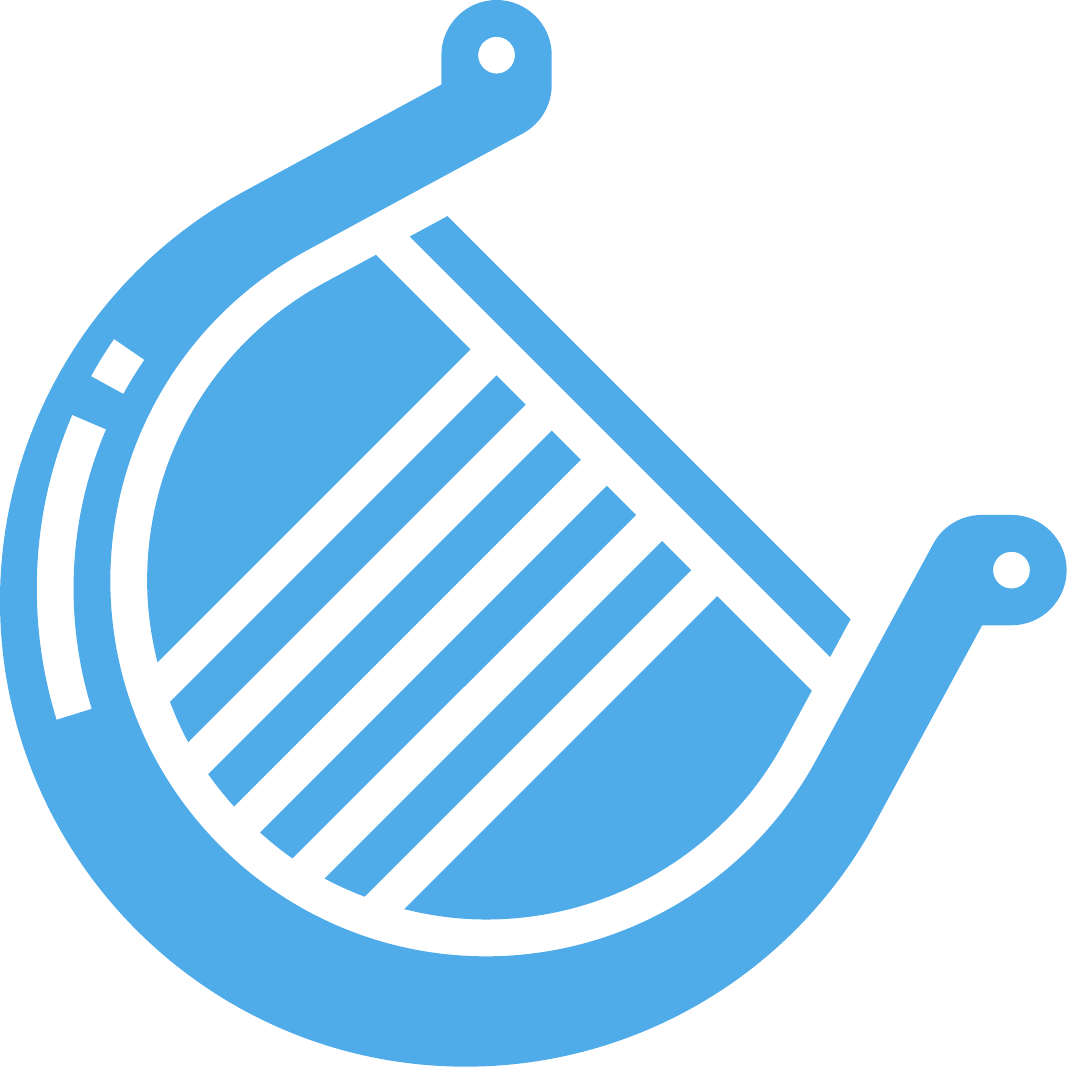}\hspace{2pt}{\color{myblue}Lyra}: An Efficient and Speech-Centric Framework for Omni-Cognition}

\newcommand*{\affmark}[1][*]{\textsuperscript{#1}}
\author{
\hspace{-30pt} {\fontsize{11.5pt}{16pt}\selectfont Zhisheng Zhong$^{1*}$ \hspace{1pt} Chengyao Wang$^{1*}$ \hspace{1pt} Yuqi Liu$^{1*}$ \hspace{1pt} Senqiao Yang$^{1}$ \hspace{1pt} Longxiang Tang$^{1}$ \hspace{1pt} Yuechen Zhang$^{1}$} \vspace{.3em}
	\\
\hspace{-36pt} {\fontsize{11.5pt}{16pt}\selectfont Jingyao Li$^{1}$ \hspace{1pt} Tianyuan Qu$^{1}$ \hspace{1pt} Yanwei Li$^{1}$ \hspace{1pt} Yukang Chen$^{1}$ \hspace{1pt} Shaozuo Yu$^{1}$ \hspace{1pt} Sitong Wu$^{1}$ \hspace{1pt} Eric Lo$^{1}$ \hspace{1pt} Shu Liu$^{2}$\affmark[\Letter] \hspace{1pt} Jiaya Jia $^{2,3}$} \vspace{.1em}
    \\
    \vspace{-10pt}
        {\hspace{-35pt} \small $^{*}$ Equal contribution \quad\quad \affmark[\Letter] Corresponding author\quad\quad Code: \url{https://github.com/dvlab-research/Lyra}} \vspace{0.9em}
	\\
	\hspace{-30pt} \small CUHK$^{1}$ \quad\quad\quad SmartMore$^{2}$ \quad\quad\quad HKUST$^{3}$
    \vspace{-4pt}
}

\twocolumn[{%
	\renewcommand\twocolumn[1][]{#1}%
	\vspace{-12mm}
	\maketitle
	\vspace{-12mm}
	\begin{center}
		\captionsetup{type=figure}
            \vspace{1.2mm}
            \centering
		\includegraphics[width=0.97\linewidth]{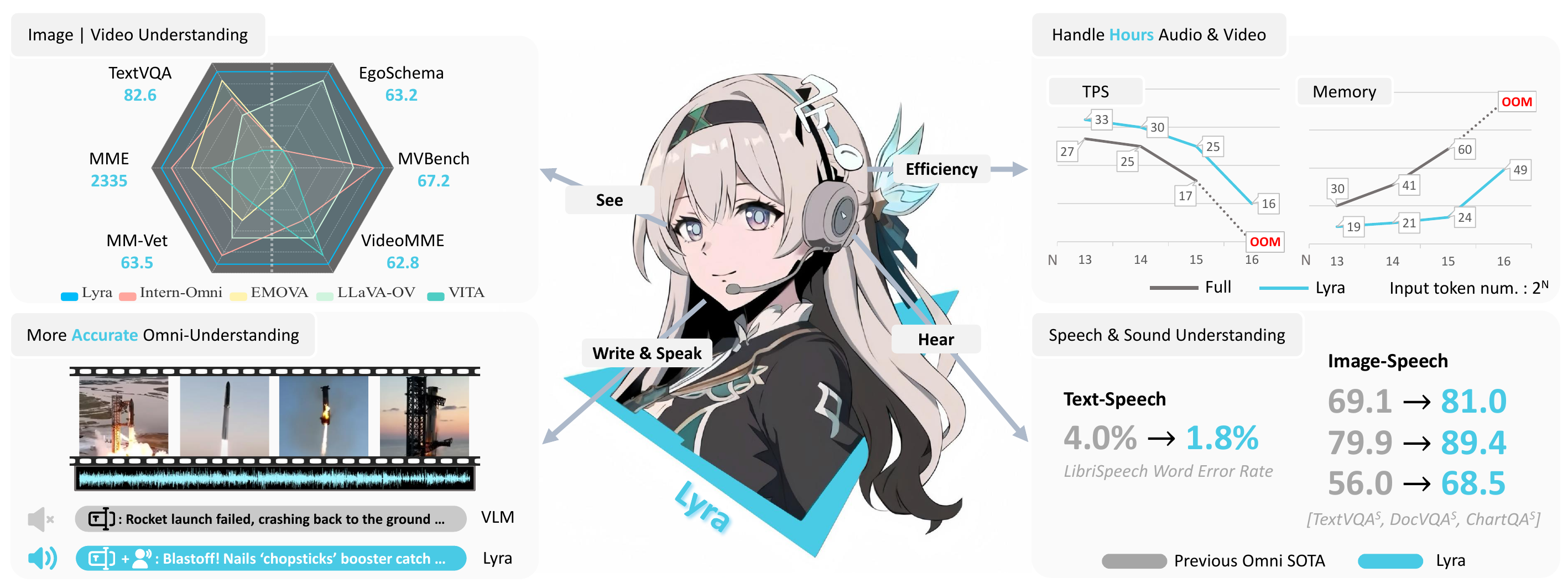}
		\vspace{-3mm}
		\captionof{figure}{\textbf{Overview of {\ours}.} {\ours} shows superiority compared with leading models in the following aspects: 1. \textit{Stronger performance.}  {\ours} achieves state-of-the-art results across a variety of modalities understanding and reasoning tasks. 2. \textit{More versatile.}  {\ours} can directly handle images, videos and audio tasks even lasting several hours. 3. \textit{More efficient.} {\ours} is trained with less data and increases the speed, reduces memory usage, making it suitable for latency-sensitive and long-context multi-modality applications.}\label{fig:teaser}
	\end{center}%
}]
\vspace{-10pt}
\begin{abstract}
As Multi-modal Large Language Models (MLLMs) evolve, expanding beyond single-domain capabilities is essential to meet the demands for more versatile and efficient AI. However, previous omni-models have insufficiently explored speech, neglecting its integration with multi-modality. We introduce Lyra, an efficient MLLM that enhances multi-modal abilities, including advanced long speech comprehension, sound understanding, cross-modality efficiency, and seamless speech interaction. To achieve efficiency and speech-centric capabilities, Lyra employs three strategies: (1) leveraging existing open-source large models and a proposed multi-moda{\textbf{\color{myblue}{l}}}it{\textbf{\color{myblue}{y}}} Lo{\textbf{\color{myblue}{RA}}} to reduce training costs and data requirements; (2) using a {\textbf{\color{myblue}{l}}}atent multi-modalit{\textbf{\color{myblue}{y}}} {\textbf{\color{myblue}{r}}}egul{\textbf{\color{myblue}{a}}}rizer and ext{\textbf{\color{myblue}{ra}}}ctor to strengthen the relationship between speech and other modalities, thereby enhancing model performance; and (3) constructing a high-quality, extensive dataset that includes 1.5M multi-modal (language, vision, audio) data samples and 12K long speech samples, enabling Lyra to handle complex long speech inputs and achieve more robust omni-cognition. Compared to other omni-methods, Lyra achieves state-of-the-art performance on various vision-language, vision-speech, and speech-language benchmarks, while also using fewer computational resources and less training data.
\vspace{-4.5mm}
\end{abstract}
   
\section{Introduction}
\label{sec:intro}

With the rapid evolution in Large Language Models (LLMs)~\cite{ChatGPT, gemma, mixtral, llama, alpaca}, empowering the impressive capabilities for multi-modality inputs is becoming an essential part of current Multimodal Large Language Models (MLLMs). However, most current MLLMs are limited to just two modalities: either vision-language~\cite{flamingo, blip2, llava, minigpt4, instructblip, ALIGN, mgm, llavaov} or speech-language~\cite{llamaomni, qwen2audio, miniomni}. OpenAI’s recent release of GPT-4o~\cite{gpt4o}, an advanced omni-modal model, has reignited interest in intelligent assistants capable of fine-grained visual perception, understanding spoken instructions, and generating vocal responses simultaneously. It highlights a strong demand for MLLMs that integrate more functions and modalities, such as visual, language, speech, sound, and even other new abilities~\cite{vita, emova, anygpt, nextgpt}. 

Based on our study, most existing omni-models~\cite{llamaomni, emova, anygpt, vita} primarily focus on the relationship between speech and text, without exploring connections between speech and other modalities, such as vision. Consequently, speech-related evaluation metrics are typically limited to text. In this paper (Sec.~\ref{sec:ab}), we observe that strong performance in the speech-text modality does not necessarily imply good performance in the speech-vision modality. Thus, we suggest that omni-model evaluation should be speech-centric, expanding its involvement with additional modalities.

To further enhance the speech capabilities of MLLMs, we inevitably encounter the following challenges: First, larger datasets (e.g., the extensive data required to train models like LLaMA3~\cite{llama3} and Qwen2-VL~\cite{qwen2vl}) are needed for both previous modalities and speech. Second, there is a clear trend toward increasing context length across modalities. More long-context benchmarks for specific modalities are being proposed, including long-document~\cite{longlora, bai2023longbench} and long-video tasks~\cite{llamavid, longvila, zhang2024long, videomme, wu2024longvideobench}. Last, building a sufficiently powerful model may demand significant financial and computational resources, which raises environmental concerns related to high carbon emissions. 

Combining the above three points, we propose {\ours}, an efficient and speech-centric framework for omni-cognition:

\vspace{2pt}
\noindent\textbf{Leveraging existing open-source large models.} We efficiently start with powerful LLMs and VLMs, like LLaMA3~\cite{llama3} and Qwen2-VL~\cite{qwen2vl}, which already demonstrate strong multi-modal capabilities. 
Through our multi-moda{\textbf{\color{myblue}l}}it{\textbf{\color{myblue}y}} Lo{\textbf{\color{myblue}RA}} module, we can effectively preserve certain strong capabilities of open-source large models in specific modalities with minimal training data, while simultaneously developing their abilities in the speech modality.

\vspace{2pt}
\noindent\textbf{Enhancing information interaction between modalities, especially within the speech modality.}
 1) Considering the implicit correspondence between speech and text, we propose {\textbf{\color{myblue}l}}atent cross-modalit{\textbf{\color{myblue}y}} {\textbf{\color{myblue}r}}egul{\textbf{\color{myblue}a}}rizer. 2) Based on instructions, we identify potential redundancy in context token information across multiple modalities. We further propose {\textbf{\color{myblue}l}}atent multi-modalit{\textbf{\color{myblue}y}} ext{\textbf{\color{myblue}ra}}ctor to mine informative tokens, which brings significant advantages in training speed, inference speed and GPU memory efficiency.

\vspace{2pt}
\noindent\textbf{High-Quality Datasets for Omni-Cognition.} Centered on speech, we have constructed two types of high-quality datasets: To enhance the model's speech capabilities, we collect and generate a multi-modal dataset of 1.5M text-image-speech samples from diverse public sources, ensuring a rich and varied data foundation; To handle longer speech inputs and demands, we are the first to construct a long speech dataset comprising 12K samples. Through training, our model achieves robust omni-cognitive abilities and can handle long speech inputs lasting several hours.

With these three improvements, {\ours} offers the following advantages (Fig.~\ref{fig:teaser}). \textbf{More versatile:} As shown in Table~\ref{tab:relate_work}, Lyra now supports both sound and speech understanding and generation, while also handling more complex long speech cases. \textbf{More efficient:} Lyra achieves faster training and inference speed across speech, image, and video tasks. Compared to previous models, Lyra has a smaller model size and is trained with less data. \textbf{Stronger:} Lyra demonstrates enhanced omni-comprehension capabilities over previous MLLMs, achieving state-of-the-art performance in vision-language and vision-speech and speech-language tasks simultaneously.

 \begin{table}[!t]
 	\centering
 	\tablestyle{2.9pt}{1.08}
 	\resizebox{0.98\linewidth}{!}{
 		\begin{tabular}{x{30pt}y{55pt}x{25pt}x{25pt}x{25pt}x{25pt}x{25pt}x{25pt}}
 			\toprule 
 			\multirow{2.5}*{Function} & \multirow{2.5}*{Method} & \multicolumn{2}{c}{Vision} & \multicolumn{4}{c}{Audio} \\
 			\cmidrule(lr){3-4} \cmidrule(lr){5-8}
 			&  & Image & Video & SU & SG & LS & Sound \\
 			\midrule
 			\multirow{3}*{Vision} & LLaVA-OV & {\mbox{\color{myblue}\cmark}} & {\mbox{\color{myblue}\cmark}} & \xmark & \xmark & \xmark & \xmark \\
 			& Intern-VL  & {\mbox{\color{myblue}\cmark}} & {\mbox{\color{myblue}\cmark}} & \xmark & \xmark & \xmark & \xmark \\
 			& Mini-Gemini & {\mbox{\color{myblue}\cmark}} & {\mbox{\color{myblue}\cmark}} & \xmark & \xmark & \xmark & \xmark \\
 			\midrule
 			\multirow{3}*{Audio} & Qwen-Audio & \xmark & \xmark & {\mbox{\color{myblue}\cmark}} & \xmark & \xmark  & {\mbox{\color{myblue}\cmark}} \\
 			& Mini-Omni & \xmark & \xmark & {\mbox{\color{myblue}\cmark}} & {\mbox{\color{myblue}\cmark}} & \xmark  & \xmark \\
 			& LLaMA-Omni & \xmark & \xmark & {\mbox{\color{myblue}\cmark}} & {\mbox{\color{myblue}\cmark}} & \xmark  & \xmark \\
 			\midrule
 			& Intern-Omni & {\mbox{\color{myblue}\cmark}} & \xmark & {\mbox{\color{myblue}\cmark}} & \xmark & \xmark  & \xmark   \\
 			& VITA & {\mbox{\color{myblue}\cmark}} & {\mbox{\color{myblue}\cmark}} & {\mbox{\color{myblue}\cmark}}  & \xmark & \xmark  & \xmark    \\
 			& Any-GPT & {\mbox{\color{myblue}\cmark}} & {\mbox{\color{myblue}\cmark}} & {\mbox{\color{myblue}\cmark}} & {\mbox{\color{myblue}\cmark}}  & \xmark   & \xmark \\
 			& EMOVA & {\mbox{\color{myblue}\cmark}} &  \xmark & {\mbox{\color{myblue}\cmark}} & {\mbox{\color{myblue}\cmark}} & \xmark  & \xmark  \\
 			\multirow{-5}*{Omni} &\textbf{{\ours}} & {\mbox{\color{myblue}\cmark}} & {\mbox{\color{myblue}\cmark}} & {\mbox{\color{myblue}\cmark}} & {\mbox{\color{myblue}\cmark}} & {\mbox{\color{myblue}\cmark}} & {\mbox{\color{myblue}\cmark}} \\
 			\bottomrule
 		\end{tabular}}
 	\vspace{-3mm}
 	\caption{\textbf{Function comparison of related work}. SU, SG, and LS represents speech understanding, speech generation,
 		and long speech support, respectively.}
 	\label{tab:relate_work}
 	\vspace{-16pt}
 \end{table}
 
 \section{Related Work}

\begin{figure*}[!t]
	\begin{center}
		\includegraphics[width=0.98\textwidth]{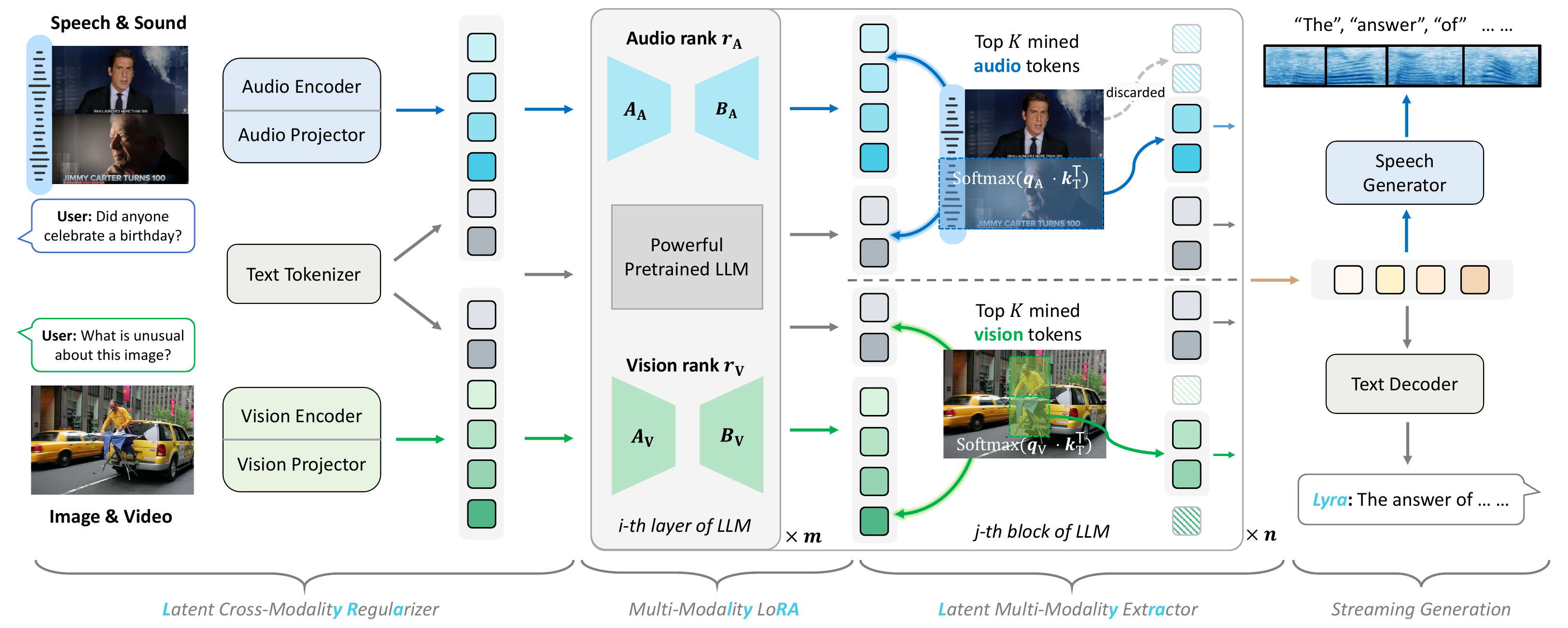}
		\vspace{-5pt}
		\caption{\textbf{The framework of {\ours}}. {\ours} supports multi-modal inputs. When the data contains a speech modality, we use the latent cross-modality regularizer to assist. Data from each modality is processed through encoders and projectors before being sent into the LLM. Within the LLM, multi-modality LoRA and latent multi-modality extraction modules operate synergistically, facilitating the simultaneous generation of both speech and text outputs. }
		\label{framework}
	\end{center}
	\vspace{-15pt}
\end{figure*}

\label{sec:relate}
\vspace{-5pt}
\textbf{Multi-modal Large Language Models.} Recent advancements in Large Language Models~(LLM) and Multimodal Large Language Models~(MLLMs) have pushed the boundaries of human-computer interaction, expanding their capabilities from text-based tasks to complex multi-modality scenarios. Large Language Models, like GPTs~\cite{ChatGPT}, LLaMA~\cite{llama, llama3} and Qwen~\cite{bai2023qwen, yang2024qwen2}, have demonstrated strong capabilities in textual understanding and generation. Building on these foundations, Vision Language Models~\cite{llava, llava1.5, llavanext, llavaov, mgm, qwen2vl, llamavid, wang2023cogvlm, lin2024vila, yao2024minicpm} extend LLMs with visual perception capabilities, leveraging advanced encoders~\cite{CLIP} and high-resolution techniques to interpret visual inputs. Speech Language Models (SLMs)~\cite{rubenstein2023audiopalm}, including SpeechGPT~\cite{zhang2023speechgpt} and LLaMA-Omni~\cite{llamaomni}, have introduced real-time speech understanding and generation, with advanced models enabling control over speech styles. Moving further, MLLMs~\cite{nextgpt} such as AnyGPT~\cite{anygpt}, VITA~\cite{vita} and EMOVA~\cite{emova}, integrate vision, text, and audio within a unified architecture, enabling robust interaction across diverse modalities. The abilities and modalities of previous leading MLLMs are listed in Table~\ref{tab:relate_work}. In contrast, {\ours} tackles complex scenarios, enabling seamless, dynamic multimodal interactions for rich, real-time AI experiences.

\vspace{1pt}
\noindent\textbf{Token Reduction for MLLMs.} Token reduction techniques aim to improve the efficiency of LLMs and VLMs by minimizing redundant tokens during inference and training. In LLMs, methods like StreamingLLM~\cite{xiao2023efficient} and FastGen~\cite{ge2023model} optimize memory usage by selectively retaining essential tokens, while techniques like H$_2$O~\cite{zhang2023h2o}, ScissorHands~\cite{liu2024scissorhands} and Quest~\cite{tang2024quest} use attention-based scoring to prioritize valuable tokens. In VLMs, approaches such as FastV~\cite{chen2025image} reduce visual tokens to tackle the high computational cost of image processing. 
{\ours} extends token reduction to more modalities, such as video and speech, where token lengths tend to increase in long-context scenarios. By evaluating the relationship between context and instruction tokens, we progressively discard redundant tokens to enhance efficiency without compromising performance.

\section{Lyra}
As shown in Fig.~\ref{framework}, the overall architecture of {\ours} is composed of four main components: latent cross-modality regularizer, multi-modality LoRA, latent multi-modality extractor, and streaming generation. {\ours} is designed as a unified framework, with each component being easily and efficiently extendable to support additional modalities and functionalities. 
In this paper, {\ours} primarily focuses on the three modalities of audio (speech, sound), vision, and language. Therefore, in the following sections of this section, We will provide a detailed introduction to the mechanisms of the following modules: latent cross-modality regularizer, multi-modality LoRA, and latent multi-modality extractor. Due to space limitations, streaming speech-text generation will be detailed in the \textit{appendix}. Since speech contexts tend to be lengthy, the integration of long speech capabilities will be discussed at the end of this section. To ensure clarity in the following discussion, let's define some key notations: the $\mathbf{X}_{[i]}$ be the token of modality-$i$. For example, $\mathbf{X}_{[\mathrm{text}]}$ represents the text token, $\mathbf{X}_{[\mathrm{image}]}$ represents the image token, $\mathbf{X}_{[\mathrm{video}]}$ represents the video token, $\mathbf{X}_{[\mathrm{speech}]}$, $\mathbf{X}_{[\mathrm{sound}]}$ represents the speech and sound token, respectively.


\subsection{{\color{myblue}L}atent Cross-Modalit{\color{myblue}y} {\color{myblue}R}egul{\color{myblue}a}rizer}

\begin{figure}[!t]
	\centering
	\includegraphics[width=0.99\linewidth]{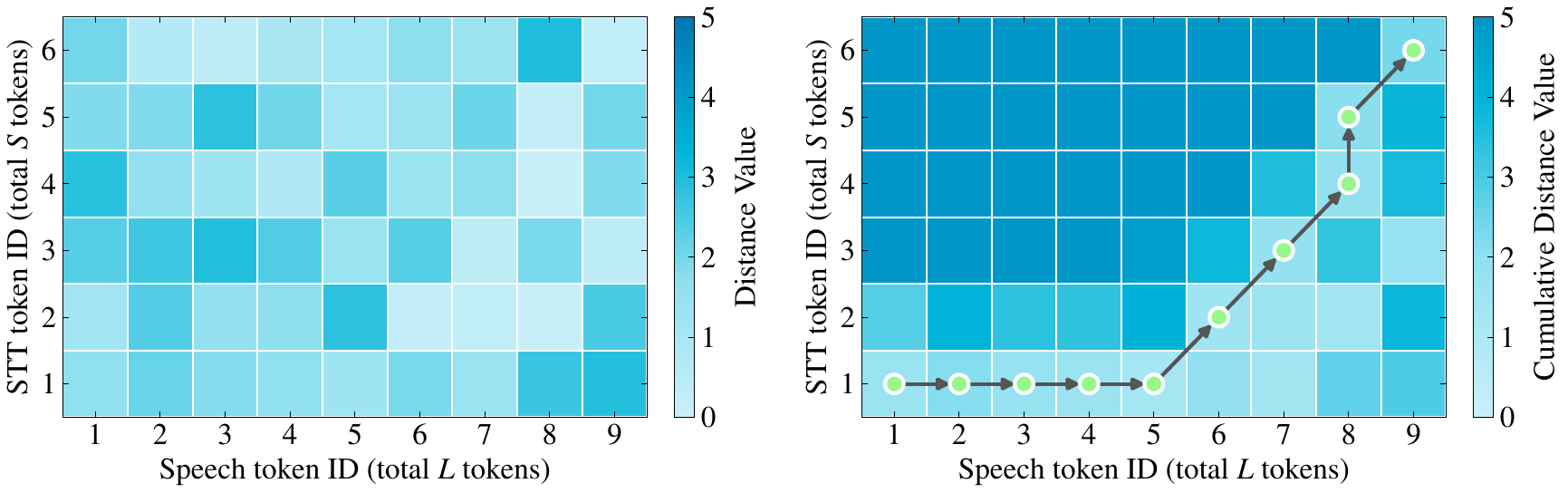}
	\caption{\textbf{Illustration of the DTW algorithm in our alignment}. Our goal is to make the speech tokens as similar as possible to the corresponding translated tokens.}
	\vspace{-10pt}
	\label{fig:dtw}
\end{figure}

For MLLMs, it is crucial to achieve effective alignment between tokens from each modality and LLM. As the view from the speech modality, there is a high degree of informational overlap with the text modality. Specifically, considering only semantic information, speech can be converted into its corresponding transcribed text. However, our experiments have shown that using speech with naive alignment training as the instruction (S+I, S for speech instruction, I for image context) generally yields less effective results compared to using transcribed text (T+I, T for text instruction, I for image context):
\begin{center}\vspace{-.2em}
	\tablestyle{4pt}{1.05}
	\resizebox{0.9\linewidth}{!}{
	\begin{tabular}{y{55pt}y{55pt}y{55pt}y{55pt}}
		TextVQA (S+I) & TextVQA (T+I) & MM-Vet (S+I) & MM-Vet (T+I) \\
		\shline
		76.7\cgaphll{-}{2.8} & \textbf{79.5} & 53.1\cgaphll{-}{8.0} & \textbf{63.1}  \\
	\end{tabular}}\vspace{-.2em}
\end{center}
To address this, we aim to make the tokens from the speech modality as similar as possible to the corresponding transcribed text tokens before feeding them into LLM, thereby minimizing the loss of relevant information. Another challenge arises from the variable length of speech: a sentence can be spoken quickly or slowly while retaining the same meaning in the text modality, leading to length discrepancies. In general, the tokens produced by a speech encoder (like Whisper) tend to be much longer than the corresponding text tokens (speech-to-text, STT), \ie, $\mathbf{X}_{[\mathrm{speech}]}\in \mathbb{R}^{d \times L}$ , $\mathbf{X}_{[\mathrm{STT}]} \in \mathbb{R}^{d \times S}$, $L > S$, $d$ is the token dimension. We define the latent distance between the $l$-th speech token and the $s$-th SST token as:
\begin{equation}
	\mathrm{dist}(l, s)\!=\!-\log\left[\mathrm{softmax}(\mathbf{X}_{\mathrm{[speech]}, l}\mathbf{X}_{\mathrm{[STT]}, s}^\top/{\tau})\right],
\end{equation}
Where $\tau$ is the temperature. To get the minimum distance between two different length tokens, we follow the Dynamic Time Warping~(DTW) algorithm:
\begin{equation}
	\mathbf{D}_{l, s} = \mathrm{dist}(l, s) + \min\{\mathbf{D}_{l, s-1}, \mathbf{D}_{l-1, s}, \mathbf{D}_{l-1, s-1}\}.
\end{equation}
The illustration is shown in Fig.~\ref{fig:dtw}. We define the latent cross-modality regularization loss as $\mathcal{L}_{\rm LCMR}=\frac{1}{L+S}\mathbf{D}_{L, S}$. Finally, the total loss of the system becomes the combination of two losses: $\mathcal{L}_{\rm total}=\mathcal{L}_{\rm CE} +\lambda\mathcal{L}_{\rm LCMR},$ where $\mathcal{L}_{\rm CE}$ is the cross-entropy loss on LLM output, and $\lambda$ is a loss weight hyper-parameter.

\subsection{Multi-Moda{\color{myblue}l}it{\color{myblue}y} Lo{\color{myblue}RA} Pipeline}
The current open-source VLM (such as Qwen2-VL) is already quite powerful. With limited data quantity and quality, jointly training vision-speech-language modalities may reduce the model's original capabilities. Therefore, we adopt an efficient multi-modality LoRA~\cite{lora} pipeline. Revisiting the notation introduced at the beginning of this section, we represent $\mathbf{X}_{[i]}$ as the token of modality-$i$. The modality-$i$ can be text, image, video, speech token, and sound. Since our model involves joint training across multiple modalities, here we define $\mathbf{X}_{[\mathrm{M}]}$ can be any combination of the above different modality tokens. The output of multi-modality LoRA can be written as:
\begin{equation}
\mathbf{H} = \left(\mathbf{B}_{[\mathrm{M}]}\mathbf{A}_{[\mathrm{M}]}+ \mathbf{W}\right)\mathbf{X}_{[\mathrm{M}]},
\end{equation}
where $\mathbf{W}$ is the original weight of LLM, $\mathbf{A}_{[\mathrm{M}]}$ and $\mathbf{B}_{[\mathrm{M}]}$ is low-rank adapter of combination-$\mathrm{M}$. During training, our Multi-Modality LoRA is integrated into each layer of the LLM. Because each modality is trained using LoRA, the process is highly efficient, achieving strong performance with minimal data while preserving much of the original model's visual capabilities.

\subsection{{\color{myblue}L}atent Multi-Modalit{\color{myblue}y} Ext{\color{myblue}ra}ctor}

As MLLMs expand their functionality and accommodate longer contexts, efficiently using tokens within a limited context window becomes essential to address the long-context problem. We now consider the relationship between non-text modalities and the text modality. In response to a given question, many tokens from non-text modalities may be largely irrelevant to the question itself. For example, as shown in Fig.\ref{framework}, only a subset of image tokens is relevant to the instructed question. Similarly, for the video and speech modality, only a portion of tokens from video and speech directly corresponds to the question instruction.

We observe that in LLM training, the long-context effect brought by high-resolution images, lengthy videos, and long audio (in the following subsection) often includes tokens with limited relevance, which not only increases the computational load for training and inference but also consumes unnecessary memory. To address this, we propose dynamically selecting multi-modality tokens based on their relevance to the text query, discarding redundant multi-modality tokens. To achieve this, we introduce a latent multi-modality information extraction strategy.

Concretely, instead of applying this strategy to every layer, we implement a block-based manner. Suppose the LLM consists of $mn$ layers; we divide them into blocks of $m$ layers each, resulting in $n$ blocks. At the final layer of each block, we apply our following information extraction strategy, which evaluates the similarity between the attention scores of tokens from each modality and the question text tokens. We represent this with the following equation:
\begin{equation}
	\mathrm{topk}\left(\mathrm{softmax}\left(\frac{\mathbf{Q}_{[\mathrm{text}]}\mathbf{K}^{\top}_{[\mathrm{\backslash text}]}}{\sqrt{d}}\right)\right),
\end{equation}
where $\mathbf{Q}_{[\mathrm{text}]}$ denotes the query corresponding to text modality tokens, and $\mathbf{K}^{\top}_{[\mathrm{\backslash text}]}$ represents the key corresponding to tokens from other modalities. For clarity, let's assume that the length of multi-modality tokens $\mathbf{K}^{\top}_{[\mathrm{\backslash text}]}$ is $L$. After passing through each block, we retain only $\rho L$ multi-modality tokens. From a block-wise perspective, the token length decays exponentially, significantly reducing computational and memory costs. A similar mechanism exists in the brain's neural processing of complex information~\cite{serences2006selective}. 
Notably, text tokens can be extended to instruction tokens for other modalities, such as speech. This extractor enables us to handle long speech more efficiently.


\subsection{{\color{myblue}L}ong Speech Capabilit{\color{myblue}y} Integ{\color{myblue}ra}tion}\label{sec:long_speech}
\begin{figure}[!t]
	\centering
	\includegraphics[width=0.99\linewidth]{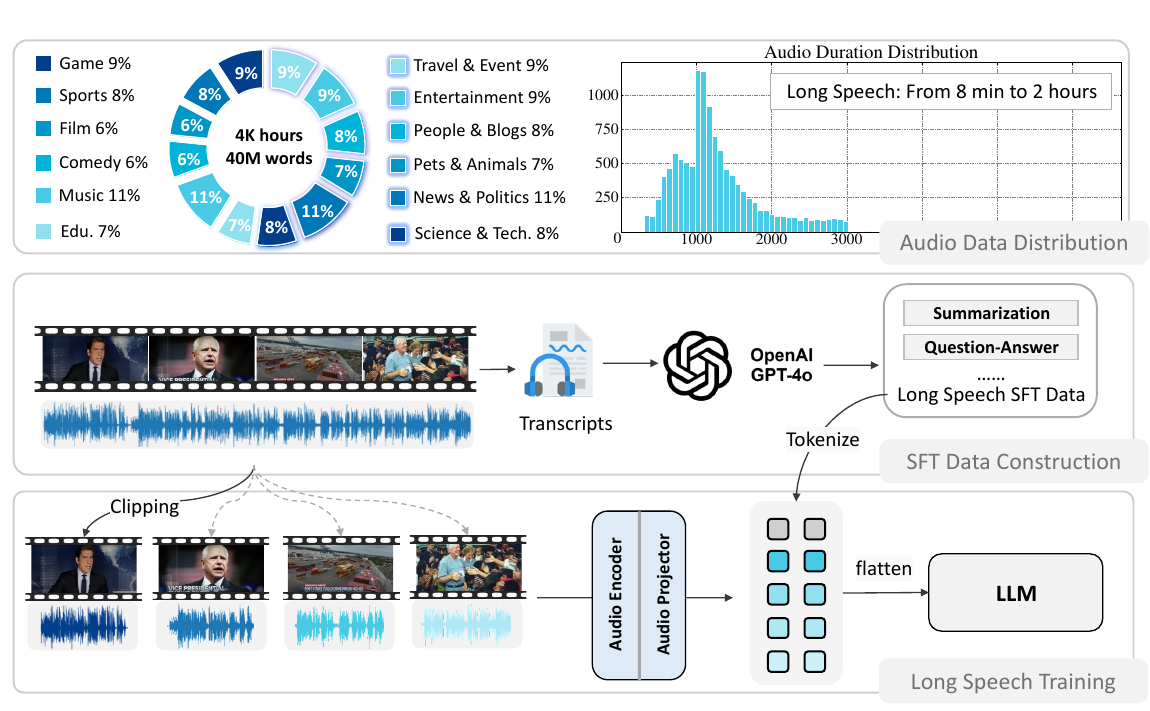}
	\caption{\textbf{Long speech capability integration pipeline}. (Middle) Our pipeline for generating instruction-following data for long speech. (Top) The proportion of question and speech categories in our long speech SFT dataset. (Bottom) Our long speech SFT pipeline. Long speech segments will be clipped and flattened.}
	\label{fig:dataset}
	\vspace{-10pt}
\end{figure}

There is a growing trend toward increasing the length of single-modality content processed by models, such as long text and long video inputs in MLLMs. However, existing MLLMs are limited in handling long speech due to the constraints of speech encoders. Specifically, models like Intern-Omni~\cite{internomni}, VITA~\cite{vita}, and LLaMA-Omni~\cite{llamaomni} use Whisper-like encoders, which restrict audio input to around 30 seconds. VITA and Mini-Omni, which employ more complex encoders, can process at most one minute of audio input. This limitation largely stems from the lack of suitable long speech SFT datasets and appropriate preprocessing methods. To address this issue, we developed the first SFT dataset for long speech understanding, aimed at enhancing model capabilities in handling extended audio content. Our dataset comprises about 12K long-form audio recordings, with durations ranging from several minutes to two hours. These recordings were collected from diverse YouTube sources, including informational videos, interviews, and speeches, covering a wide range of topics—from humanities and current events to technology and society. With related transcripts, we utilized LLM to generate question-and-answer pairs derived from the captions and instructions. These questions cover summarization and other types of inquiries that encourage a comprehensive understanding of long speech content. The overall question distribution and details are illustrated in Fig.~\ref{fig:dataset}.

Once the dataset was ready, we tackled the challenge with the speech encoder. Inspired by high-resolution image segmentation methods like LLaVA-NeXT~\cite{llavanext}, we adopted a similar strategy to better handle the speech encoder for long audio processing (illustrated in Fig.~\ref{fig:dataset}). However, unlike previous speech cases, a new challenge emerged: for a naive Whisper-v3 encoder, a 30-second audio clip is encoded into 1,500 tokens. Under typical short speech scenarios, an LLM can handle 1,500 tokens comfortably. When we consider long speech cases, such as a two-hour audio clip, this would result in an astonishing 360,000 tokens, which is beyond our processing capacity. Thus, it is essential to consider compression techniques on speech tokens. The experimental results are presented as follows:
\begin{center}\vspace{-.2em}
	\tablestyle{2.9pt}{1.08}
	\resizebox{0.95\linewidth}{!}{
		\begin{tabular}{y{45pt}y{32pt}y{32pt}y{32pt}y{32pt}y{32pt}}
			\#(Token) & 100& 150 & \cellcolor[HTML]{efefef} 300 & 500 & 1500\\
			\shline
			TextVQA$^{\rm S}$ &75.9\% & 76.8\% & \cellcolor[HTML]{efefef} 77.8\% & 78.0\% & 76.8\% \\
            MM-Vet$^{\rm S}$ &55.3 & 54.4\% & \cellcolor[HTML]{efefef} 56.3\% & 58.8\% & 58.9\% \\
	\end{tabular}}\vspace{-.2em}
\end{center}
Experimental results indicate that having a higher number of speech tokens provides certain benefits. However, beyond a certain threshold, the performance improvement becomes quite limited. Taking into account both computational costs and model performance, we ultimately decided to use the 300 compressed tokens version for extending the model to handle long speech cases. 

\section{Experiments}

In this section, \textit{we conduct a speech-centric evaluation, assessing its integration with image, video, and text modalities}. we first outline our experimental framework, commencing with the experimental setup.
Subsequently, we compare {\ours} with leading methods on various benchmarks and qualitative results. Detailed component wise analysis (\textit{based on {\ours}-Base}) is given at the end of this section. More experiment details and results refer to our \textit{Appendix}.

\subsection{Experimental Setup}

\noindent\textbf{Implementation Details.} In this study, we instantiate {\ours} with the following designs and settings: 

\noindent 1. Strong vision encoders and LLMs: Building on the previously applied vision model Qwen2-VL's ViTs and LLMs~\cite{qwen2vl}, they can now process images of any resolution, dynamically converting them into a variable number of visual tokens. We have also designed three versions: For {\ours}-Mini, we use Qwen2-VL 2B. For {\ours}-Base, we apply Qwen2-VL 7B. For {\ours}-Pro, we choose Qwen2-VL 72B.

\noindent 2. Efficient audio encoder: We adopted Whisper-large-v3~\cite{radford2023robust} ({\ours}-Base and {\ours}-Pro) and its light-weight version, Whisper-large-v3-turbo ({\ours}-Mini), which have been trained on a large amount of audio data and has strong capabilities in speech recognition and translation.

\noindent 3. Four stage training for omni-cognition (refer to our appendix for specific details): In the first stage, we conduct text-to-speech pretraining to train the speech encoder. In the second stage, we perform joint training on text, image, and speech modalities to train the LLM along with the corresponding projectors. In the third stage, we train the LLM to extend the model’s capability in handling long speech. In the fourth stage, we train our speech generator, enabling the model to simultaneously output text and corresponding audio in a streaming manner.

\vspace{5pt}
\noindent\textbf{Datasets and Evaluations.} For model optimization, we construct high-quality data for omni-understanding and speech generation. 

\noindent 1. High-quality multi-modal dataset: Based on the Mini-Gemini SFT~\cite{mgm} dataset, we carefully collected and extended a high-quality multi-modal dataset that covers common scenes and document images and speeches. It contains about 1.5M open-source image-speech, text-image, and text-speech instruction samples. To enhance the generalization of speech modality, we utilize ChatTTS~\cite{ChatTTS} with varying configurations to generate different audios. 

\noindent 2. Long speech SFT dataset: As mentioned in Sec.~\ref{sec:long_speech}, we constructed a delicate long speech SFT dataset for long speech capability integration with 12K samples. The dataset involves a distribution of longer audio durations and covers a wide range of domains.

\noindent 3. Evaluation: Unlike the previous omni-model~\cite{vita, emova}, which only tested text-to-speech capabilities, we employed a more omni comprehensive evaluation that covers interactions across image, video, text, and speech modalities.

\begin{table*}[!t]
	\begin{center}
		\centering
		\tablestyle{2.9pt}{1.08}
		\scalebox{0.9}{
			\begin{tabular}{y{60pt}|y{28pt}x{39pt}x{39pt}x{39pt}x{39pt}x{39pt}x{39pt}x{39pt}x{39pt}x{39pt}x{40pt}}
				\toprule 
				\multicolumn{2}{l}{Omni Comparison} & \multicolumn{3}{c}{Text-Image} & \multicolumn{3}{c}{Text-Video} & \multicolumn{3}{c}{Image-Speech} & Text-Speech \\
				\cmidrule(lr){3-5} \cmidrule(lr){6-8} \cmidrule(lr){9-11} \cmidrule(lr){12-12}
				Method & Params.  & TextVQA & MME & MM-Vet & VideoMME & MVBench & Egoschema &  TextVQA$^{\rm S}$  & DocVQA$^{\rm S}$ & ChartQA$^{\rm S}$ & LibriSpeech$\downarrow$\\
				\midrule
				Mini-Gemini  & 8B & 71.9 & 1989 & 53.5 & - & - & - & - & - & - & - \\
				LLaVA-OV & 7B & 65.4 & 1998 & 57.5 & 58.2 & 56.7 & 60.1 & - & - & - & - \\
				Intern-VL2  & 8B & 77.4 & 2211 & 60.0 & 54.0 & 66.4 & - & - & - & - & - \\
				\midrule
				Mini-Omni & 7B & - & - & - & - & -  & - & - & - & - & 4.5 \\
				SALMONN & 13B & - & - & - & - & -  & - & - & - & - & 2.1 \\
				Qwen2-Audio  & 8B & - & - & - & - & -  & - & - & - & - & {1.6} \\
				\midrule
				Intern-Omni & 8B & 80.6 & 2210 & 60.0 & - & -  & - & 69.1 & 79.9 & 56.0 & - \\
				VITA & 66B & - & 2097 & 41.6  & 59.2 & -  & -  & - & - & - & 8.1 \\
				EMOVA & 14B & 82.0 & 2205 & 55.8 & - & -  & - & - & - & - & 4.0 \\
				\midrule
				\cellcolor[HTML]{efefef} \textbf{{\ours-Mini}} & \cellcolor[HTML]{efefef}  3B & \cellcolor[HTML]{efefef}  78.3 & \cellcolor[HTML]{efefef} 1884 & \cellcolor[HTML]{efefef} 51.2 & \cellcolor[HTML]{efefef} 55.0 & \cellcolor[HTML]{efefef} 62.5 & \cellcolor[HTML]{efefef} 54.1 & \cellcolor[HTML]{efefef} 73.4 & \cellcolor[HTML]{efefef} 74.8 & \cellcolor[HTML]{efefef} 40.7 & \cellcolor[HTML]{efefef} 2.1 \\
				\cellcolor[HTML]{efefef} \textbf{{\ours-Base}} & \cellcolor[HTML]{efefef}  9B & \cellcolor[HTML]{efefef}  {82.6} & \cellcolor[HTML]{efefef}  {2335} & \cellcolor[HTML]{efefef}  {63.5} & \cellcolor[HTML]{efefef}  {62.8}& \cellcolor[HTML]{efefef}  {67.2} & \cellcolor[HTML]{efefef} {63.2} & \cellcolor[HTML]{efefef} {80.0} & \cellcolor[HTML]{efefef} {85.5} & \cellcolor[HTML]{efefef} {61.0} & \cellcolor[HTML]{efefef} 2.0 \\
				\cellcolor[HTML]{efefef} \textbf{{\ours-Pro}} & \cellcolor[HTML]{efefef}  74B & \cellcolor[HTML]{efefef}  83.5 & \cellcolor[HTML]{efefef}  2485 & \cellcolor[HTML]{efefef}  71.4 & \cellcolor[HTML]{efefef}  69.9 & \cellcolor[HTML]{efefef}  72.3 & \cellcolor[HTML]{efefef}  75.8 & \cellcolor[HTML]{efefef} 81.0 & \cellcolor[HTML]{efefef} 89.4 & \cellcolor[HTML]{efefef} 68.5 & \cellcolor[HTML]{efefef} 1.8\\
				\bottomrule
			\end{tabular}
		}
		\vspace{-7pt}
		\caption{\textbf{Omni-comparison on vision-language-speech benchmarks}. Bench$^{\rm S}$ indicates that it uses speech instruction as the input.}
		\label{tab:main}
	\end{center}
	\vspace{-18pt}
\end{table*}

\subsection{Main Results}

\noindent\textbf{Quantitative Results.} In the quantitative analysis experiments, we primarily compare our model with current leading VLMs, such as Mini-Gemini~\cite{mgm}, Llava-OV~\cite{llavaov}, Intern-VL2~\cite{chen2024far}, and SLM, like Mini-Omni~\cite{miniomni}, SALMONN~\cite{tangsalmonn}, Qwen2-Audio~\cite{qwen2audio}, and Omni models including Intern-Omni~\cite{internomni}, AnyGPT~\cite{anygpt}, VITA~\cite{vita}, and EMOVA~\cite{emova}. The input modalities we compare are also the most widely used, including text-image, text-video, image-speech, and text-speech. Detailed results are presented in Table~\ref{tab:main}.
In calculating the total parameters of the model, we considered all modality-specific encoders, projectors, and related components. Our model includes three versions: a mini version (3B), a based version (9B), and a pro version (74B). Benefiting from multi-modality LoRA and Qwen2-VL, our model maintains relatively high performance in text-image and text-video tasks. For the speech modality, as we mentioned in Introduction part, previous models have evaluated the speech modality rather crudely, without extensively testing metrics for interactions between the speech modality and other modalities. Our model comprehensively outperforms existing omni models in both image-speech (with an improvement of approximately 9\%) and text-speech (with an improvement of approximately 2\%) tasks. Additionally, our model is more lightweight, requiring fewer training samples.

\vspace{5pt}
\noindent\textbf{Qualitative Results.} To ascertain the omni comprehension prowess of {\ours} in real world settings, we apply it to a variety of understanding and reasoning tasks in the bottom left part of Fig.~\ref{fig:teaser} and our \textit{Appendix}. By contrast, {\ours} can well solve more complex multi-modality cases.


\subsection{Component-Wise Analysis}\label{sec:ab}

\vspace{5pt}
\noindent\textbf{Latent Cross-Modality Regularizer.} We first delve into the proposed latent cross-modality regularizer and report results in Table~\ref{tab:alignment}. It is clear that the model achieves significant gains for both speech-image inputs and text-image inputs, with the regularizer integrated as an assistance between speech modality and text modality. In the training of the image-speech-text tri-modal model, introducing the $\mathcal{L}_{\rm LCMR}$ significantly enhances the performance of both image-speech and image-text alignments, reducing the gap between them. We also observe that, with only $\mathcal{L}_{\rm CE}$, image-text performance lags behind image-speech by 8\% on the MM-Vet benchmark. However, the performance of speech-text remains relatively unchanged whether using the CE loss or joint loss. Therefore, previous omni models~\cite{emova, vita} that assessed the speech modality just based on the LibriSpeech~\cite{panayotov2015librispeech} WER metric for speech-text alignment are rather arbitrary. We need to evaluate the performance of the speech modality alongside other modalities to accurately measure the effectiveness of omni-models. This also demonstrates the effectiveness of our $\mathcal{L}_{\rm LCMR}$.

\vspace{5pt}
\noindent\textbf{Latent Multi-Modality Extractor.} For the latent multi-modality extractor~(LMME) module, we focus primarily on its \textbf{efficiency} and \textbf{effectiveness} in multi-modal tasks. First, we analyze its efficiency, with specific results summarized in Tables~\ref{tab:info_min_infer} and \ref{tab:info_min_train}. In Table~\ref{tab:info_min_infer}, we vary the token length, ranging from \(2^{11}\) to \(2^{17}\) (under a long-context case). We denote LMME($n$, $\rho$) as splitting the LLM into $n$ blocks, with each block retaining the top $\rho$ proportion of the most important tokens. We compare three models: the baseline, LMME(4, 0.8), and LMME(4, 0.7). The key metrics examined include Prefill Time, tokens-per-second (TPS), and memory usage on the A100 GPUs. Under the baseline model, multimodal content exceeding \(2^{15}\) tokens results in out-of-memory~(OOM) errors. In contrast, our models LMME(4, 0.8) and LMME(4, 0.7) still have room for \(2^{17}\) tokens, consuming over 50\% less memory. Additionally, the Prefill Time is significantly shorter than the baseline model (by 100\%), and the token generation speed is also notably faster (by 50\%).

In Table~\ref{tab:info_min_train}, we primarily examine the improvement in training speed. We evaluate it using our proposed Lyra SFT and long-speech SFT dataset, which contains 1.5M samples and 12K samples, respectively. From the table, our LMME can reduce training time by more than 50\% compared to the original. Since the context in the long-speech dataset is generally longer than it in the 1.5M dataset, the acceleration effect becomes even more pronounced. 

\begin{figure*}[!t]
	\begin{center}
		\includegraphics[width=0.95\textwidth]{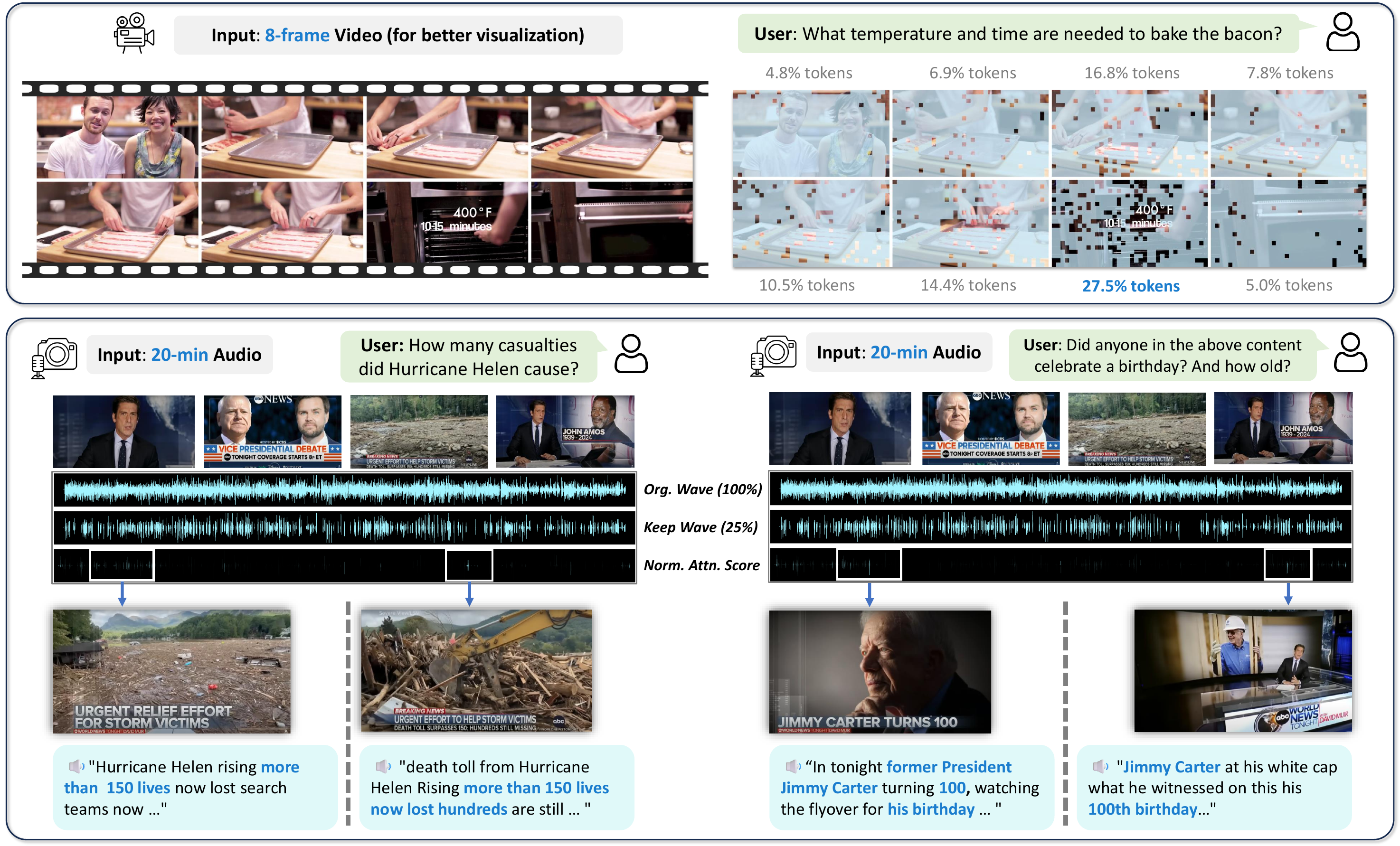}
		\vspace{-5pt}
		\caption{\textbf{Visualization of latent multi-modality extractor in various modalities.} The upper part is the video modality, and the lower part is the audio modality. Through latent multi-modality information extraction, semantic tokens related to the instruction are retained, reducing the computational cost of the MLLM. The visualization of the image modality and different blocks can be found in the \textit{appendix}.}
		\label{fig:visualization}
		\vspace{-12pt}
	\end{center}
\end{figure*}

To verify the effectiveness of our extractor module, we examine the retention of multi-modal tokens. We primarily assess three types of tokens: image tokens, video tokens, and speech tokens. The specific visualizations are shown in Fig.~\ref{fig:visualization}. As seen in the figure, our model ultimately retains only about 10\%-25\% of the tokens across all three modalities. Moreover, the retained token positions are highly relevant to the user-provided instructions, effectively helping to remove information unrelated to the instructions and thereby accelerating training and inference. We also have included the performance experiments related to LMME in the \textit{appendix} section.

\begin{table}[!t]
	\centering
	\tablestyle{2.9pt}{1.08}
	\resizebox{0.98\linewidth}{!}{
		\begin{tabular}{y{60pt}x{30pt}x{30pt}x{30pt}x{30pt}x{40pt}}
			\toprule
			Effectiveness & \multicolumn{2}{c}{TexVQA} & \multicolumn{2}{c}{MM-Vet} & LibriSpeech \\
			\cmidrule(lr){2-3} \cmidrule(lr){4-5}\cmidrule(lr){6-6}
			Type & S+I & T+I & S+I & T+I & S+T \\
			\midrule
			Baseline & - & \textbf{82.3} & - & \textbf{62.8} & - \\
			$\mathcal{L}_{\rm CE}$  & 76.7 & 79.5 & 53.1 & 61.1 & \cellcolor[HTML]{efefef} \textbf{1.9} \\
			\cellcolor[HTML]{efefef} $\mathcal{L}_{\rm CE}$ + $\lambda \mathcal{L}_{\rm LCMR}$ & \cellcolor[HTML]{efefef} \textbf{77.8} & \cellcolor[HTML]{efefef} 80.1 & \cellcolor[HTML]{efefef} \textbf{58.1} & \cellcolor[HTML]{efefef} 62.6 & \cellcolor[HTML]{efefef} 2.0 \\
			\bottomrule
	\end{tabular}}
	\vspace{-1mm}
	\caption{\textbf{Latent cross-modality regularizer}. With our regularizer, the performance of both the speech-image and text-image modalities improves, and the gap narrows.}
	\label{tab:alignment}
	\vspace{-10pt}
\end{table}
\begin{table}[!t]
	\begin{minipage}{\linewidth}
		\centering
		\tablestyle{2.9pt}{1.15}
		\resizebox{0.98\linewidth}{!}{
			\begin{tabular}{llx{20pt}x{20pt}x{20pt}x{20pt}x{20pt}x{20pt}x{20pt}}
				\toprule
				Metric & \# (Tokens) & $2^{11}$ & $2^{12}$ & $2^{13}$ & $2^{14}$ & $2^{15}$ & $2^{16}$ & $2^{17}$ \\
				\midrule
				\multirow{3}*{Prefill(s)$\downarrow$} & Baseline  & 0.19 & 0.33 & 0.65 & 1.47 & 2.99 & \cellcolor[HTML]{efefef}\textbf{OOM} & \cellcolor[HTML]{efefef}\textbf{OOM} \\
				& LMME(4, 0.8) & 0.17 & 0.24 & 0.44 & 0.76 & 1.60 & 4.24 & 10.2 \\
				&LMME(4, 0.7) & 0.16 & 0.21 & 0.37 & 0.59 & 1.23 & 3.05 & 7.75 \\
				\midrule
				\multirow{3}*{TPS$\uparrow$}  & Baseline & 32.6 & 30.8 & 27.3 & 25.3 & 16.6 & \cellcolor[HTML]{efefef}\textbf{OOM} & \cellcolor[HTML]{efefef}\textbf{OOM} \\
				& LMME(4, 0.8) & 32.7 & 31.5 & 31.8 & 28.6 & 22.7 & 14.1 & 8.37 \\
				& LMME(4, 0.7) & 33.8 & 33.3 & 32.5 & 30.1 & 25.3 & 16.6 & 10.1 \\
				\midrule
				\multirow{3}*{Memory$\downarrow$} & Baseline & 20G & 23G & 30G & 41G & 60G & \cellcolor[HTML]{efefef}\textbf{OOM} & \cellcolor[HTML]{efefef}\textbf{OOM} \\
				& LMME(4, 0.8) & 17G & 18G & 19G & 21G & 24G & 33G & 49G \\
				& LMME(4, 0.7) & 17G & 18G & 19G & 21G & 24G & 33G & 49G \\
				\bottomrule
		\end{tabular}}
		\vspace{4pt}
		\begin{subfigure}[t]{\linewidth}
			\caption{Prefill time,  tokens per second (TPS), GPU memory comparison.}
			\label{tab:info_min_infer}
		\end{subfigure}
		\vspace{-2pt}
	\end{minipage} \\
	\begin{minipage}{\linewidth}
		\centering
		\tablestyle{2.9pt}{1.08}
		\resizebox{0.98\linewidth}{!}{
			\begin{tabular}{y{74pt}x{38pt}x{48pt}x{48pt}x{48pt}}
				\toprule
				Data Type & Baseline & LMME(4, 0.9) & LMME(4, 0.8) & LMME(4, 0.7) \\
				\midrule
				{\ours}-MM-1.5M & 66h & 58h\ \ \cgaphll{-}{13\%} & 47h\ \ \cgaphll{-}{29\%} & 41h\ \ \cgaphll{-}{38\%}  \\
{\ours}-LongSpeech-12K & 9.6h & 7.0h\ \cgaphll{-}{27\%} & 5.7h\ \cgaphll{-}{40\%} & 4.5h\ \cgaphll{-}{54\%}  \\
				\bottomrule	
		\end{tabular}}
		\vspace{4pt}
		\begin{subfigure}[t]{\linewidth}
			\caption{Training time on multi-modality datasets comparison.}
			\label{tab:info_min_train}
		\end{subfigure}
		\vspace{-2pt}
		\vspace{-10pt}
		\caption{\textbf{Efficiency of latent multi-modality extractor}.}
		\vspace{-10pt}
	\end{minipage} 
\end{table}

\begin{figure*}[!t]
	\centering
	\begin{minipage}{0.68\textwidth}  
		\centering
		\includegraphics[width=\textwidth]{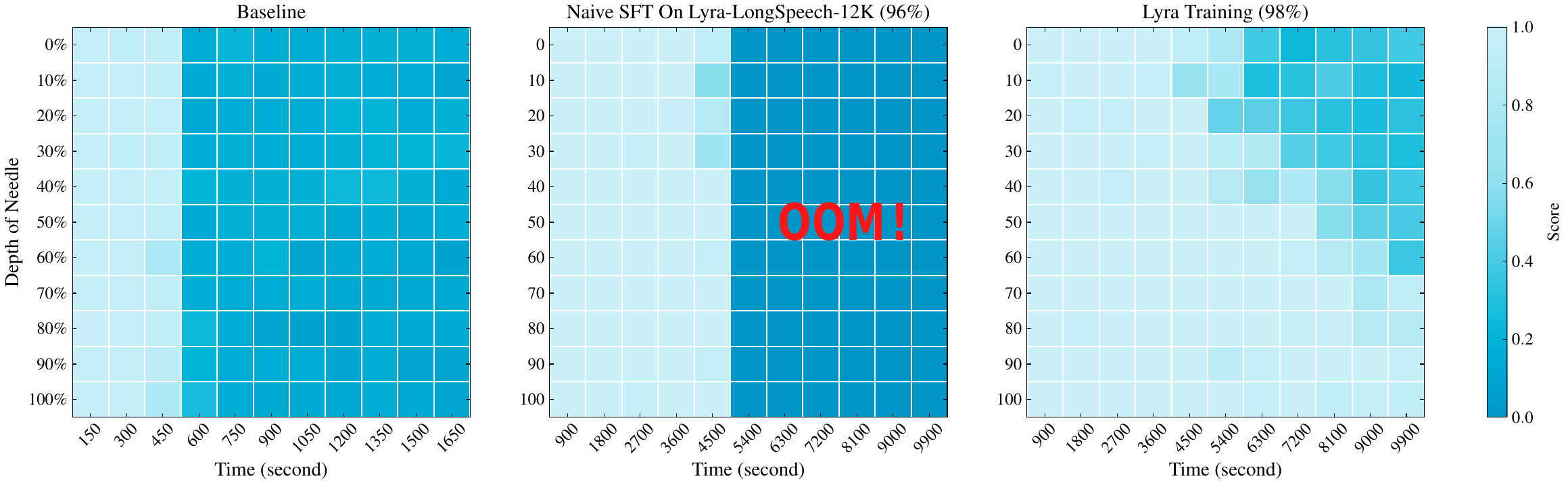}
	\end{minipage}%
	\begin{minipage}{0.28\textwidth}  
		\centering
		\includegraphics[width=\textwidth]{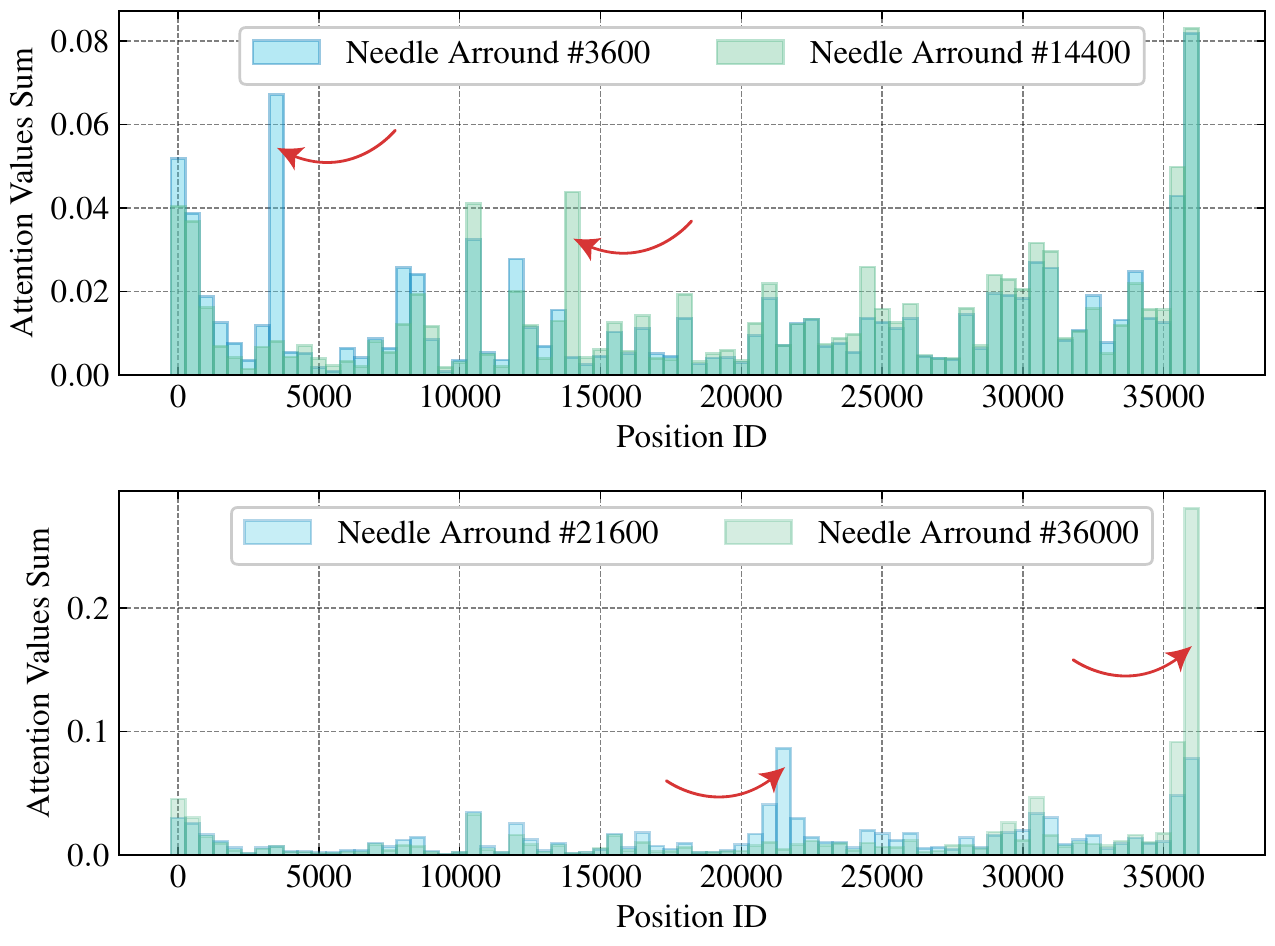}
	\end{minipage}\\
	\begin{subfigure}[t]{0.27\textwidth}
		\caption{}
		\label{fig:needle_base}
	\end{subfigure}
	\begin{subfigure}[t]{0.13\textwidth}
		\caption{}
		\label{fig:needle_sft}
	\end{subfigure}
	\begin{subfigure}[t]{0.28\textwidth}
		\caption{}
		\label{fig:needle_lyra}
	\end{subfigure}
	\hfill
	\begin{subfigure}[t]{0.3\textwidth}
		\caption{}
		\label{fig:needle_attn}
	\end{subfigure}
	\\
	\vspace{-5pt}
	\caption{\textbf{Comparison of needle in long speech haystack (average with five samples)}. (a) The baseline
		model can not retrieve right needles after \textbf{450} seconds. (b) Model finetuned on our long speech datasets can not retrieve right needles after \textbf{4,500} seconds and achieves 96\% accuracy in 4,500 seconds.  (c) Our latent extractor, trained on our long speech datasets, can retrieve longer audio (\textbf{9,900} seconds), and presents 98\% accuracy in 4,500 seconds. (d) As the position of the needle changes, the attention in our model also shifts accordingly.}
	\label{fig:needle}
	\vspace{-10pt}
\end{figure*}

\vspace{5pt}
\noindent\textbf{Long-Speech Capability Integration.}
After performing SFT on our Lyra long speech 12K data mentioned Sec.~\ref{sec:long_speech}, we design the following experiments to validate the model's capabilities in processing long speech and latent multi-modality extraction, given the current lack of a long-speech benchmark. The first experiment is the long speech ``Needle in a Haystack'' evaluation. We selected five audio files, each more than 3 hours in length, and inserted open-ended audio questions and answers at various points throughout the files. The results are shown on the left side of Fig.~\ref{fig:needle}. According to the figure, we observe that, without enhancing long-speech processing capabilities, the model can handle up to approximately eight minutes of audio. beyond that length, it fails to generate a proper output (Fig.~\ref{fig:needle_base}). However, with SFT on our Lyra long speech 12K data, the model can handle audio lengths of up to 4,500 seconds. With audio exceeding 4,500 seconds, the model’s memory usage surpasses the limit (Fig.~\ref{fig:needle_sft}). By leveraging the latent multi-modality extractor module, we achieve the ability to process even longer audio, extending up to and beyond two hours (Fig.~\ref{fig:needle_lyra}). Additionally, In Fig.~\ref{fig:needle_attn}, we visualize the token-level attention retention and variations for the ``needle'' with the information extractor module, under the same question instructions. Notably, we can see that as the needle is placed in different locations, the information extractor module dynamically adjusts the attention distribution and retention for positions accordingly.

\begin{table}[!t]
	\begin{minipage}{\linewidth}
		\centering
		\tablestyle{2.9pt}{1.08}
		\resizebox{0.98\linewidth}{!}{
			\begin{tabular}{y{74pt}x{40pt}x{44pt}x{44pt}x{44pt}}
				\toprule
				Method & Overall & Short & Medium & Long \\
				\midrule
				Baseline (\textbf{7B}) & 62.8 & 73.8& 62.3 & 52.3  \\
				Baseline + subtitle & 64.4 & 76.2 & 63.4 & 53.4  \\
				\cellcolor[HTML]{efefef} LSCI (\textbf{7B}, solve 33\%) & \cellcolor[HTML]{efefef} \textbf{78.6} & \cellcolor[HTML]{efefef} \textbf{89.8} & \cellcolor[HTML]{efefef} \textbf{77.7} & \cellcolor[HTML]{efefef} \textbf{74.8}  \\
				Baseline + LSCI &  66.2 &  75.7 &  64.0 &  58.9 \\
				\midrule
				GPT-4o~\cite{gpt4o} + subtitle & 77.1 & 82.8 & 76.6 & 72.1  \\
				\bottomrule	
		\end{tabular}}
		\vspace{-1mm}
		\caption{\textbf{Effectiveness of long speech capability integration}. {\ours}  integrated with long speech ability, using only audio input, can handle one-third of VideoMME cases, and its accuracies on long, medium, short metrics are better than the current best VLM.}
		\vspace{2mm}
		\label{tab:long_speech}
	\end{minipage} \\
	\begin{minipage}{\linewidth}
		\centering
		\tablestyle{2.9pt}{1.08}
		\resizebox{0.98\linewidth}{!}{
			\begin{tabular}{y{40pt}y{60pt}x{40pt}x{40pt}x{40pt}}
				\toprule 
				Modality & Benchmark & Baseline & + SFT &  + \cellcolor[HTML]{efefef} MLoRA  \\
				\midrule
				\multirow{3}*{Image} & TextVQA~\cite{textvqa} & 82.3 & 81.3 & \cellcolor[HTML]{efefef} \textbf{82.6}\\
				& MME~\cite{mme}  & 2332 & 2275 & \cellcolor[HTML]{efefef} \textbf{2335} \\
				& MMMU~\cite{mmmu}  & 49.2 & 48.7 & \cellcolor[HTML]{efefef} \textbf{50.8} \\
				\midrule
				\multirow{3}*{Video} & VideoMME~\cite{videomme} & 62.8 & 61.0 & \cellcolor[HTML]{efefef} \textbf{62.8}  \\
				& MVBench~\cite{mvbench} & 66.7 & 66.8 & \cellcolor[HTML]{efefef} \textbf{67.2}   \\
				& EgoSchema~\cite{ego} & 62.4 & \textbf{63.5} & \cellcolor[HTML]{efefef} 63.2  \\
				\midrule
				& TextVQA$^{\rm S}$~\cite{textvqa} & - & 77.8 & \cellcolor[HTML]{efefef} \textbf{80.0} \\
				& DocVQA$^{\rm S}$~\cite{docvqa} & - & 84.0 & \cellcolor[HTML]{efefef} \textbf{84.6}  \\
				\multirow{-3}*{Speech} & MM-Vet$^{\rm S}$~\cite{mmvet}  &- &  54.0  & \cellcolor[HTML]{efefef} \textbf{60.0} \\
				\bottomrule
			\end{tabular}
		}
		\vspace{-1mm}
		\caption{\textbf{Effectiveness of multi-modality LoRA~(MLoRA)}. For powerful pretrained models, adding new modality can impair the abilities of other modalities. MLoRA can effectively address it.}
		\label{tab:mlora}
		\vspace{-3mm}
	\end{minipage}
\end{table}


The second experiment is based on VideoMME. This benchmark includes videos ranging from 30 seconds to one hour. We first extract the audio from these videos and feed only the audio data into our long speech model to obtain predictions and perform the VideoMME evaluation. Along with generating predictions, we also require our model to output whether it can answer the question based on the audio alone. Specific results are shown in Table~\ref{tab:long_speech}. From the table, it is evident that long audio can resolve about one-third of the test samples, with model accuracy exceeding 78\%, significantly outperforming the 7B model. We integrate the long-speech output into our Lyra model, which ultimately performs better than using subtitles alone.

\vspace{5pt}
\noindent\textbf{Multi-Modality LoRA~(MLoRA) Pipeline.}
The effectiveness results of MLoRA are presented in Table~\ref{tab:mlora}. Compared to multi-modal SFT, MLoRA maintains better original vision performance while enhancing the capability in new modalities like speech. Additionally, our framework is more efficient, achieving better results with less data (50\%).
\begin{center}\vspace{-.2em}
	\tablestyle{2.9pt}{1.08}
	\resizebox{0.9\linewidth}{!}{
		\begin{tabular}{y{50pt}y{50pt}y{50pt}y{50pt}}
			Intern-Omni & VITA& EMOVA & \cellcolor[HTML]{efefef} {\ours}\\
			\shline
			27M samples & 5M samples & 4M samples & \cellcolor[HTML]{efefef} \textbf{2.7M} samples\\
	\end{tabular}}
\end{center}

\section{Conclusion}
In conclusion, Lyra represents a significant step forward in MLLMs, \textit{efficiently} integrating complex speech, vision, and language modalities with reduced computational requirements (\textit{less data, faster speed}). We focus on speech to enhance its interaction with other modalities within MLLMs. By leveraging the proposed modules, and high-quality, comprehensive SFT datasets, Lyra achieves state-of-the-art performance across vision-speech,  speech-language, and vision-language benchmarks, which is a more comprehensive evaluation for omni-models to previous research. Our experiments also reveal that \textit{speech plays a critical role} in multimodal understanding, yet current MLLMs do not effectively leverage this information. We hope our work encourages future researchers to further explore and harness the potential of speech/long speech within MLLMs.

\clearpage	

\clearpage
\clearpage
\onecolumn
\appendix

{\centering
\Large
\textbf{\thetitle}\\
\vspace{0.5em}Supplementary Material \\
\vspace{1.0em}} 

\etoctoccontentsline{part}{Supplementary Material}


{\color{myblue}\textbf{We strongly recommend that readers watch the video in our supplementary materials, which include more audio and video examples to get a better understanding and experience.}} In the following supplementary material,  we provide more details about the training configurations and the construction and information of our dataset in Sec.~\ref{sec:config}. In Sec.~\ref{sec:more_results}, we present additional module settings along with some experimental results and analyses. In Sec.~\ref{sec:quan}, we showcase the qualitative results of {\ours}.

\setlength{\cftbeforesecskip}{0.5em}
\cftsetindents{section}{0em}{1.8em}
\cftsetindents{subsection}{1em}{2.5em}
\cftsetindents{subsubsection}{3.0em}{3.5em}
\renewcommand{\cftdotsep}{0.5}


\renewcommand{\cftdotsep}{0.5}

\renewcommand{\contentsname}{Appendix Contents}


\section{Training Configuration and Data}\label{sec:config}
\subsection{Detailed Training Configuration}
\noindent\textbf{Stage-1: Speech Alignment.} In this stage, we only train the parameters of the speech projector for speech-language
pre-alignment with the LibriSpeech~\cite{panayotov2015librispeech} and Common Voice Corpus~\cite{CommonVoice} datasets, with about 1.0M data samples.

\vspace{5pt}
\noindent\textbf{Stage-2: Joint Text-Image-Speech Training.} Based on the Mini-Gemini~\cite{mgm} SFT data, we assemble and construct a unified dataset with 1.5M samples for the image-text-speech joint training. We use the ChatTTS~\cite{ChatTTS} model to convert high-quality SFT data from text instructions into speech instructions. The multi-modal dataset, \ie, {\ours}-MultiModal-1.5M, includes not only single-turn instructions but also multi-turn instructions.

\vspace{5pt}
\noindent\textbf{Stage-3: Long Speech SFT.} To enable the model to integrate the long speech capability, we construct the first long-speech SFT dataset, called {\ours}-LongSpeech-12K. Details can be found in Sec.~\ref{sec:long_speech} of the main paper. To ensure more robust performance, the dataset covers a wide range of topics, including humanities, social sciences, technology, education, and more. At this stage, we train both the speech module and the whole LLM module.

\vspace{5pt}
\noindent\textbf{Stage-4: Streaming Text-Speech Generation.} During the speech generation stage, we only train the speech generator. To better align the speech generator with the text decoder, we exclusively use text-speech modality QA pairs in our dataset. We filtered and selected a portion of suitable data from the datasets in our Stage-1, Stage-2, and Stage-3 for speech generation, resulting in a dataset of approximately 227K samples.

\noindent Detailed training settings are further explicated in Table~\ref{tab:training}.

\begin{table*}[!h]
\begin{center}
    \centering
    \tablestyle{2.9pt}{1.2}
    \scalebox{0.95}{
      \begin{tabular}{y{10pt}y{60pt}|x{100pt}|x{100pt}|x{100pt}|x{100pt}}
  \toprule
  & \textbf{Settings} & \textbf{Stage-1} & \textbf{Stage-2} & \textbf{Stage-3} & \textbf{Stage-4} \\
  \midrule
  \multirow{2}{*}{\rotatebox[origin=c]{90}{\footnotesize \textit{Speech}}} & \textbf{Audio Length} & $<$ 30s   & $<$ 30s & $<$ 2500s, 30s clips & $<$ 30s \\
  & \# Tokens & $300$ & $300$   & Max $25,000$  & $300$  \\
  \midrule
  \multirow{2}{*}{\rotatebox[origin=c]{90}{\footnotesize \textit{Data}}} & \textbf{Dataset} & LibriSpeech + CommonVoice & {\ours}-MultiModal-1.5M & {\ours}-LongSpeech-12K & Filter from Stage-1, 2, 3\\
  & \# Samples & 1.2M & 1.5M & 12K & 227K \\
  \midrule
  \multirow{4}{*}{\rotatebox[origin=c]{90}{\footnotesize \textit{Training}}} & \textbf{Trainable} & Projector & Projector + LLM & Projector + LLM & Speech Generator \\
  & \textbf{Batch Size} & 256 & 128 & 16 & 32 \\
  & \textbf{Learning rate} & $1\times 10^{-3}$ & $2\times 10^{-4}$ & $2\times 10^{-4}$ & $2\times 10^{-4}$\\
  & \textbf{Epoch} & 1 & 1 & 3 & 1\\
  \bottomrule
  \end{tabular}
    }
    \vspace{-7pt}
    \caption{\textbf{Detailed training settings of {\ours}}.}
    \label{tab:training}
\end{center}
\vspace{-18pt}
\end{table*}

\subsection{Data Collection and Curation}
To ensure the data quality and training efficiency, we consider the following aspects while generating speech data for three modalities of joint training.

\vspace{5pt}
\noindent\textbf{Generate multi-modal interleave data.} To ensure models' ability to process interleaved multi-modal data, we randomly select one round from multi-round conversations and convert its text into speech, while keeping the remaining rounds in text format. This guarantees that our SFT data preserves its multi-modal interleaved structure.

\vspace{5pt}
\noindent\textbf{Oral Expression.} Certain types of text are not well-suited for direct conversion using TTS technology. In these cases, we ensure the content is rewritten in a more conversational, oral form. For example, we rephrase ``A:" as ``Option A is" to enhance clarity and naturalness.

\vspace{5pt}
\noindent\textbf{Speaker Diversity.} To maintain diversity in our generated speech, we randomly select speakers with varying timbres and pitches for each instance. 
Since ChatTTS~\cite{ChatTTS} obtains different speaker characteristics through various Gaussian sampling, it exhibits great diversity and robustness. During our generation process, we switch to a new set of ChatTTS random samples every 128 instructions.

\vspace{5pt}
\noindent\textbf{Be Aware of the OCR Text.} In real-world applications, a MLLM retrieves text by calling the OCR interface, such as TextVQA. Many OCR tokens, such as `G0' and `EF', lack clear meaning and are not suitable for verbal expression as speech input. Following this practice, we do not convert OCR text into speech.

\noindent Here, we list some training prompts and evaluation examples of our data in Table~\ref{tab:data}.

\section{More Component-Wise Details \& Analysis}\label{sec:more_results}




\subsection{{\color{myblue}L}atent Multi-Modalit{\color{myblue}y} Ext{\color{myblue}ra}ctor}

\begin{figure*}[!t]
	\begin{center}
		\includegraphics[width=0.99\textwidth]{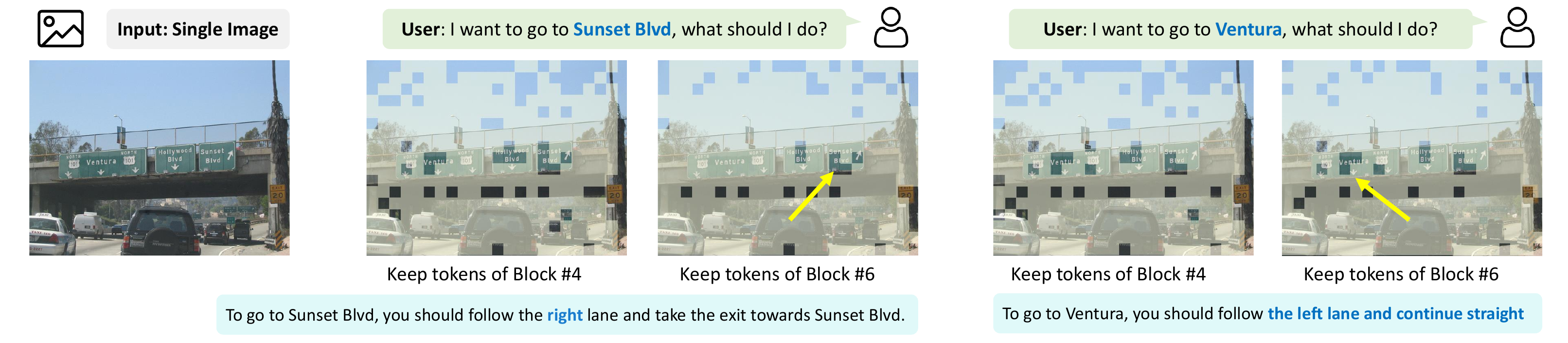}
		\vspace{-5pt}
		\caption{\textbf{Visualization of latent multi-modality extractor in the image modality}. }
		\label{fig:visualization_img}
		\vspace{-12pt}
	\end{center}
\end{figure*}

Qwen2-VL is exceptionally powerful, with the quantity and quality of its training data far surpassing those of public datasets and open-source models. As a result, most approaches to continual learning based on Qwen2-VL tend to result in performance degradation. Therefore, to evaluate the performance of our extractor module, we opt to train a new model from scratch. The results are shown in Table~\ref{tab:extractor}. Under the same training settings, models using latent multi-modality extractor achieve faster training speeds, with a maximum acceleration of nearly 50\%. Additionally, they maintain or even improve average performance by up to 1\% across multiple benchmarks. This series of experiments demonstrates the effectiveness of our extractor. Visualization of the latent multi-modality extractor in image modality is shown in Fig.~\ref{fig:visualization_img}. From it, the tokens retained in different blocks are all related to the user's instruction. Additionally, for different questions, the token regions in the image most relevant to the question are preserved. This result is consistent with the video and speech modalities discussed in our main paper.

\begin{table*}[!t]
	\begin{center}
		\centering
		\tablestyle{2.9pt}{1.08}
		\scalebox{0.95}{
			\begin{tabular}{y{40pt}|x{45pt}y{45pt}y{53pt}y{33pt}x{30pt}x{30pt}x{30pt}x{33pt}x{30pt}x{30pt}y{50pt}}
				\toprule 
				Method & LLM & Vision  & Data & Time & TextVQA & MME & MM-Vet & MMB-EN & SEED & MMMU & Avg. Rate \\
				\midrule
				Baseline &Vicuna-7B          & CLIP+Conv        & {\ours}-MM-1.5M            & 65h           & 68.4             & 1865         & 41.3            & 65.8            & 68.1          & 36.8          &      100.0\%          \\
               + Extractor &Vicuna-7B          & CLIP+Conv        & {\ours}-MM-1.5M           & 35h\cgaphll{-}{46\%}           & 69.9             & 1899         & 44.9            & 66.7            & 67.5          & 35.3          & 101.5\%\cgaphl{+}{1.5\%}        \\ \midrule
Baseline &Qwen2-7B           & SigLIP               & LLaVA-665K          & 18h           & 69.7            & 1974         & 39.4            & 76.7            & 74.2         & 40.8          &   100.0\%             \\
+ Extractor &Qwen2-7B              & SigLIP               & LLaVA-665K          & 14h\cgaphll{-}{22\%}           & 69.1            & 2005         & 38.6            & 76.9            & 73.5         & 40.6          & 99.6\%\ \ \cgaphll{-}{0.4\%}         \\ \midrule
Baseline &Qwen2-7B               & SigLIP               & {\ours}-MM-1.5M           & 51h           & 71.9            & 2030         & 51.0              & 78.1            & 74.5         & 40.2          &    100.0\%           \\ 
+ Extractor &Qwen2-7B               & SigLIP               & {\ours}-MM-1.5M           & 35h\cgaphll{-}{31\%}           & 71.8            & 2007         & 50.6            & 77.7            & 73.7         & 42.1          & 100.1\%\cgaphl{+}{0.1\%}          \\ \bottomrule

			\end{tabular}
		}
		\vspace{-7pt}
		\caption{\textbf{Latent multi-modality extractor training performance}. The training time is reduced by an average of \textbf{one-third}, while the average performance does not degrade and even \textbf{improves by 0.4\%}.}
		\label{tab:extractor}
	\end{center}
	\vspace{-10pt}
\end{table*}

\begin{figure}[t!]
\centering
\tcbset{
  colback=gray!5!white, 
  colframe=myblue,       
  width=0.95\linewidth,     
  boxrule=1pt,          
  arc=4mm,              
  left=5pt,             
  right=5pt,            
  top=5pt,              
  bottom=5pt,           
}
\begin{tcolorbox}[title={\ours} Data Examples]
\footnotesize 
\textbf{Training conversations}:
\vspace{2pt}

human: \texttt{<image>\textbackslash n}What are the two people holding?\texttt{\textbackslash n}Answer the question using a single word or phrase.
\vspace{2pt}

GPT: Umbrella.
\vspace{2pt}

human: What is the person with the Red Hat doing?
\vspace{2pt}

GPT: Taking pictures.
\vspace{2pt}

human: \texttt{<speech>}
\vspace{2pt}

GPT: Blanket.
\vspace{2pt}

\vspace{2pt}
\textbf{Evaluation cases}:
\vspace{1pt}

human: \texttt{<image>\textbackslash n}Reference OCR token: DAKOTA, DIGITAL, Single-Use, Camera, Pire, digitat\texttt{\textbackslash n<speech>}
\normalsize
\end{tcolorbox}
\vspace{-10pt}
\caption{\textbf{{\ours} training and evaluation data examples.}}
\label{tab:data}
\end{figure}

\subsection{{\color{myblue}L}ong Speech Capabilit{\color{myblue}y} Integ{\color{myblue}ra}tion}

In this part, we primarily introduce prompts related to the long speech capability. The detailed prompts are shown in Table~\ref{table:prompt}. The first is the GPT-4o-based prompt used to generate Q\&A during the long speech data collection process. The second is the inference prompt we used to apply the long-speech Lyra model on the VideoMME benchmark. For detailed results and analysis, refer to Sec.~\ref{sec:long_speech} and the long-speech capability integration part in Sec.~\ref{sec:ab}.

\begin{figure*}[t!]
\centering
\tcbset{
  colback=gray!5!white, 
  colframe=myblue,       
  width=0.95\textwidth,     
  boxrule=1pt,          
  arc=4mm,              
  left=5pt,             
  right=5pt,            
  top=5pt,              
  bottom=5pt,           
}
\begin{tcolorbox}[title=Long Speech Question-Answer Generation Prompt Example]
\footnotesize 
\textbf{Task:}
\vspace{2pt}

You will be provided with a transcript from an audio or video recording. Your task is to generate question-answer pairs based on the content of the transcript.
\vspace{1pt}

Guidelines for Question-Answer Pair Generation:
\vspace{1pt}

- The first question should be about summarizing the content of this recording.
\vspace{1pt}

- Carefully read the transcript provided and base all questions and answers strictly on the content within.
\vspace{1pt}

- Ensure that each question is directly related to specific details in the transcript, such as events, facts, or points made by the speaker.
\vspace{1pt}

- Provide clear, concise, and specific questions, along with accurate answers derived from the transcript.
\vspace{1pt}

- Do not introduce any new information that isn't in the transcript. If the speaker does not introduce themselves, refer to them as ``Speaker'' or ``Narrator''.
\vspace{1pt}

- Avoid generic or overly broad questions; aim for a range of question types (e.g., factual, inferential, explanation-based).
\vspace{1pt}

- Generate five question-answer pairs.
\vspace{1pt}

\vspace{3pt}
\textbf{Output Format:}
\vspace{2pt}

- Your output should be structured as a JSON object.
\vspace{1pt}

- Each question-answer pair should be formatted as:
\vspace{1pt}






\texttt{```}json
\vspace{-\baselineskip}
\begin{verbatim}
{
    [
        {"Question": <question-1>, "Answer": <answer-1>},
        {"Question": <question-2>, "Answer": <answer-2>},
        ...
    ]
}
\end{verbatim}
\vspace{-\baselineskip}
\texttt{```}
\normalsize
\end{tcolorbox}

\begin{tcolorbox}[title=Long Speech VideoMME Evaluation Prompt Example]
\footnotesize 
Based on the context, determine if it provides enough information to answer the question: 
\vspace{1pt}

\texttt{<question>} with the provided choices
\texttt{<option-A>}, \texttt{<option-B>}, \texttt{<option-C>}, \texttt{<option-D>}. 
\vspace{1pt}

Do not introduce any information not found in the context.
\vspace{1pt}

- If the context is sufficient to answer the question, respond ``yes'' and answer with the option's letter from the given choices directly.

- If the context does not contain enough information to answer the question, respond ``no''.
\normalsize
\end{tcolorbox}
\vspace{-10pt}
\caption{\textbf{Long speech related prompt examples.}}
\label{table:prompt}
\end{figure*}

\subsection{Sound Capabi{\color{myblue}l}it{\color{myblue}y} Integ{\color{myblue}ra}tion}

For the sound modality, due to the lack of many pretrained models, we primarily follow ImageBind\cite{imagebind} as the sound encoder. ImageBind processes sound, text, and image modalities using a training approach similar to CLIP~\cite{CLIP}, ultimately encoding them into just one single token. This approach is not particularly generalizable. During the sound SFT process, our model based on LLaMA3~\cite{llama3} is trained on the AudioCaps~\cite{kim2019audiocaps} dataset, which contains a total of 46K training samples. The quantitative performance of our model on the test set is shown in Table~\ref{tab:sound}. 

\begin{figure*}[!t]
	\centering
	\includegraphics[width=0.97\linewidth]{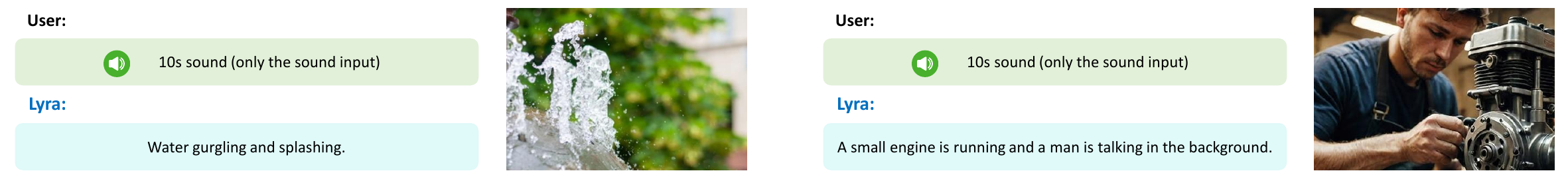}
	\caption{\textbf{Sound capability qualitative results.}}
	\vspace{-5pt}
	\label{fig:sound}
\end{figure*}

Regarding this dataset, as the authors of AudioCaps~\cite{kim2019audiocaps} have noted, ``Even to humans, recognizing the true identity of a sound can be ambiguous." Moreover, LLM-based multimodal models tend to produce more detailed descriptions, while metrics like SPICE~\cite{anderson2016spice} and CIDEr~\cite{vedantam2015cider} are outdated and fail to effectively reflect the most suitable results. Even under such circumstances, our {\ours}, trained on just 46K samples for the sound modality, outperforms previous sound models. Some qualitative results are shown in Fig.~\ref{fig:sound}.

\subsection{Streaming Text-Speech Generation}
For the speech-text streaming generation component, we primarily refer to LLaMA-Omni~\cite{llamaomni} to enable the MLLM to output speech audio.

\vspace{5pt}
\noindent\textbf{Speech Discretization.}
To handle speech responses, we discretize the audio into discrete units with the following steps:
1). Continuous representations are extracted using the HuBERT model~\cite{hsu2021hubert}.
2). These representations are clustered into discrete indices via the K-means algorithm.
3). Consecutive repeated indices are merged to form a sequence of discrete units, which can be converted back to waveforms using a vocoder~\cite{polyak21_interspeech}.

\vspace{5pt}
\noindent\textbf{Speech Decoder for Streaming Generation.}
A streaming speech decoder is introduced after the LLM to enable simultaneous generation of text and speech:
To ensure the overall structure remains consistent with the LLM, the decoder is built using two transformer layers similar to Qwen2-VL~\cite{qwen2vl}. Similar to LLaMA-Omni, it processes the hidden states from the LLM and generates discrete speech units in a non-autoregressive manner~\cite{zhang2024streamspeech, ma2024non}. For upsampling, the text hidden states from the LLM are upsampled to match the speech sequence's length. These upsampled representations are processed by the speech decoder to produce output features for the discrete speech units.

\vspace{5pt}
\noindent\textbf{Alignment and CTC Training.} Following LLaMA-Omni, Connectionist Temporal Classification (CTC)~\cite{ctc} is used to align the decoder’s output with the discrete speech units. During training, the model learns to match the output features to the target speech units by minimizing the CTC loss. During inference, the most likely sequence is selected, converted into discrete units, and passed through the vocoder to generate audio.

\subsection{TTS Methods Ablation Study}


In this subsection, we briefly compare the impact of different TTS (text-to-speech) methods on the generalization and robustness of speech instruction (across different domains). We primarily used two TTS methods: ChatTTS~\cite{ChatTTS} and Edge-TTS~\cite{edgetts}. ChatTTS employs Gaussian sampling to simulate different speakers (As shown in Listing~\ref{lst:sample_random}), while Edge-TTS randomly selects from a fixed set of 41 speakers. ChatTTS is likely to be more diverse. We trained models using instruction data generated by these TTS methods and evaluated TextVQA speech instructions generated by different TTS methods. Detailed results can be found in Table~\ref{tab:tts1}. Models trained with speech generated by ChatTTS demonstrated better generalization due to its diversity.

Similar results were observed when compared with speech instructions generated by Intern-Omni~\cite{internomni}. Because we cannot access their training speech instruction data; they only provided the evaluation speech instruction data of DocVQA and ChartQA. Specific results are provided in Table~\ref{tab:tts2} and \ref{tab:tts3}. While models perform better when trained and evaluated on instructions generated by the same system, the experiments overall demonstrate that instructions generated by ChatTTS are more robust compared to the other two methods.

\section{Qualitative Results}\label{sec:quan}

\subsection{Examples of Images and Videos}
\begin{figure*}[!t]
	\begin{center}
		\includegraphics[width=0.9\textwidth]{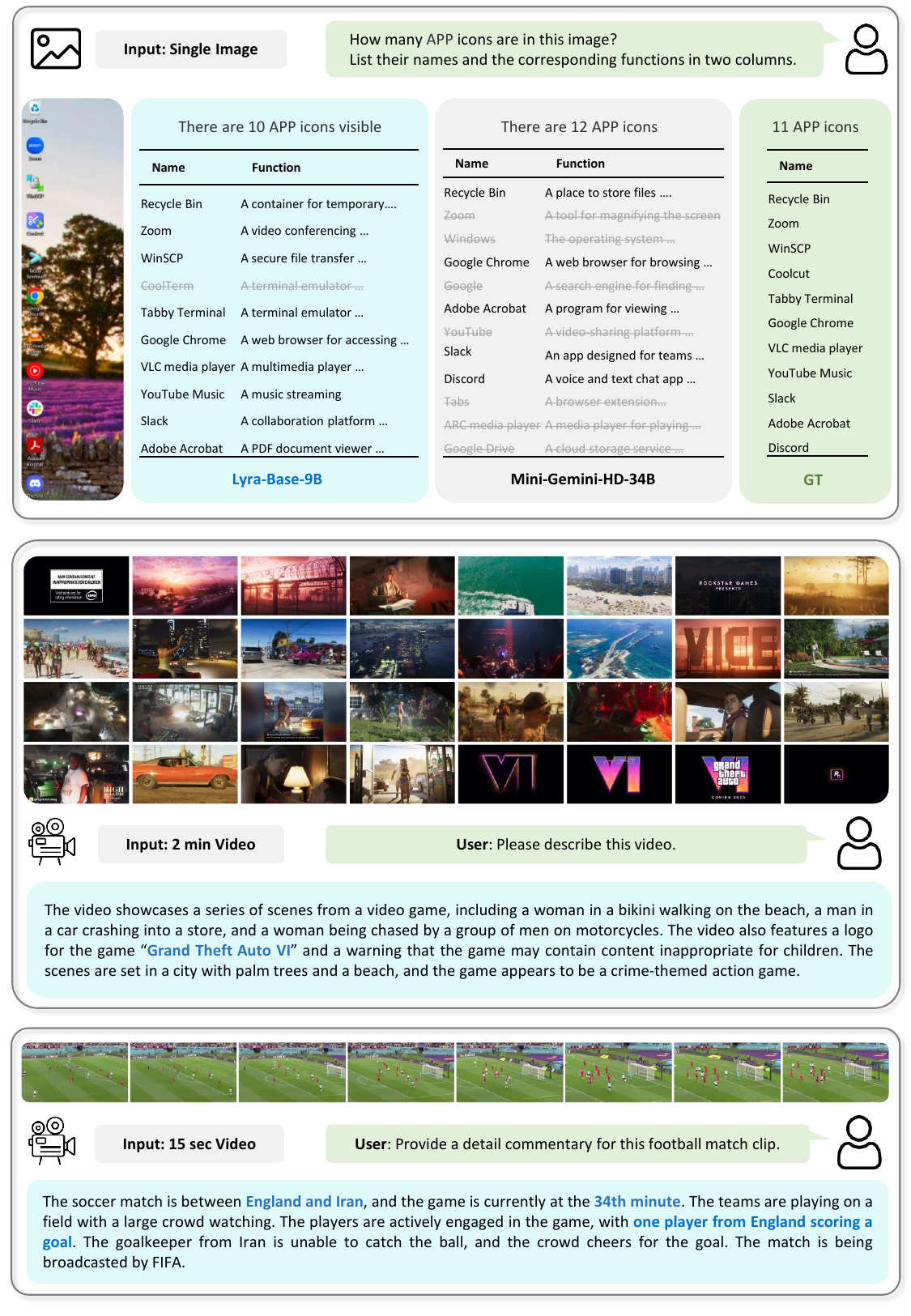}
		\vspace{-5pt}
		\caption{Image-text and video-text qualitative results of {\ours}.}
		\label{fig:image_video}
	\end{center}
\end{figure*}

In Fig.~\ref{fig:image_video}, we present additional
interactions with {\ours}, showcasing the model's
adeptness in knowledge-based perception and reasoning for both images and videos. In various complex scenarios, such as recognition of complex PC backgrounds, understanding of game interfaces, and analyzing football match videos with significant differences between frames, {\ours} demonstrates superior understanding and reasoning cognitive outcomes.

\begin{table*}[t!]
    \begin{minipage}{0.47\linewidth}
	\centering
	\tablestyle{2.9pt}{2.0}
	\scalebox{0.9}{
		\begin{tabular}{x{45pt}x{45pt}x{45pt}x{45pt}x{45pt}x{45pt}}
				\toprule
				AT~\cite{mei2021audio} & BART~\cite{gontier2021automated} & PairMix~\cite{kim2022exploring} & CoDi~\cite{tang2024any} & \cellcolor[HTML]{efefef}  {\ours}-Base  \\
				\midrule
				 16.8 & 17.7 & 18.1 & 17.1 &  \cellcolor[HTML]{efefef}  19.5 \\
				\bottomrule
			\end{tabular}}
	\caption{\textbf{Sound SPICE performance comparison.}}
        \vspace{-5pt}
        \label{tab:sound}
    \end{minipage}
    \hspace{5pt}
    \begin{minipage}{0.52\linewidth}
    \begin{subfigure}{0.4\linewidth}  
    \centering
    \tablestyle{2.9pt}{1.08}
    \scalebox{0.8}{
    \begin{tabular}{x{38pt}|x{28pt}x{34pt}}
    \toprule
    {Eval/Train} &  ChatTTS& Edge-TTS\\
    \midrule
    ChatTTS     & \cellcolor[HTML]{efefef} \textbf{80.0} & \cellcolor[HTML]{efefef} \textbf{79.5}\\
    Edge-TTS     & 79.7& 78.3 \\
    \bottomrule
    \end{tabular}}
    \caption{TextVQA$^{\rm S}$}
    \label{tab:tts1}
    \end{subfigure}%
    \hfill
    \begin{subfigure}{0.28\linewidth}  
    \centering
    \tablestyle{2.9pt}{1.08}
    \scalebox{0.8}{
    \begin{tabular}{x{38pt}|x{28pt}}
    \toprule
    {Eval/Train} &  ChatTTS\\
    \midrule
    ChatTTS     & 84.6 \\
    Intern-O    & 82.3 \\
    \bottomrule
    \end{tabular}}
    \caption{DocVQA$^{\rm S}$}
    \label{tab:tts2}
    \end{subfigure}%
    \hfill
    \begin{subfigure}{0.28\linewidth}  
    \centering
    \tablestyle{2.9pt}{1.08}
    \scalebox{0.8}{
    \begin{tabular}{x{38pt}|x{28pt}}
    \toprule
    {Eval/Train} &  ChatTTS\\
    \midrule
    ChatTTS     & 60.4 \\
    Intern-O     & 58.3 \\
    \bottomrule
    \end{tabular}}
    \caption{ChartQA$^{\rm S}$}
    \label{tab:tts3}
    \end{subfigure}
    \vspace{-5pt}
    \caption{\textbf{Different TTS training and evaluation.}}
    \vspace{-10pt}
    \end{minipage}
\end{table*}

\lstset{
    basicstyle=\ttfamily\footnotesize,
    keywordstyle=\color{blue}\bfseries,
    commentstyle=\color{gray},
    stringstyle=\color{red},
    numbers=left,
    numberstyle=\tiny\color{gray},
    stepnumber=1,
    frame=single,
    breaklines=true,
    tabsize=4,
    language=Python
}
\begin{table}[!t]
\begin{lstlisting}[language=Python, caption=Sample Random Function in ChatTTS (Pytorch),  label=lst:sample_random]
def sample_random(self) -> torch.Tensor:
    spk = (
        torch.randn(self.dim, device=self.std.device, dtype=self.std.dtype)
        .mul_(self.std)
        .add_(self.mean)
    )
    return spk
\end{lstlisting}
\end{table}

\subsection{Examples of Long Speeches} 

\begin{figure*}[!t]
	\begin{center}
		\includegraphics[width=0.9\textwidth]{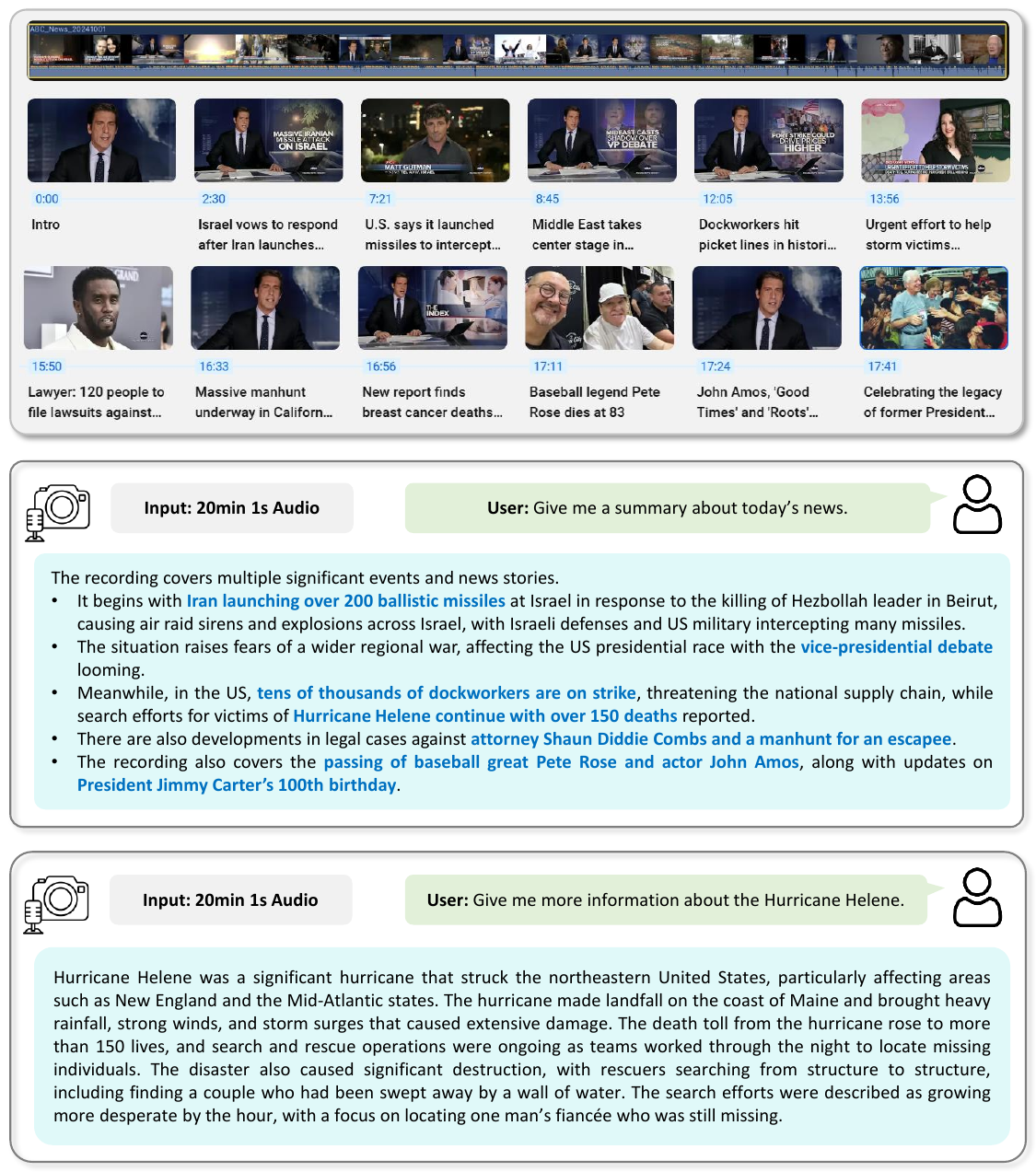}
		\vspace{-5pt}
		\caption{{{\ours} long speech capability qualitative results for handling daily news}.}
		\label{fig:long_speech1}
		\vspace{-12pt}
	\end{center}
\end{figure*}

\begin{figure*}[!t]
	\begin{center}
		\includegraphics[width=0.9\textwidth]{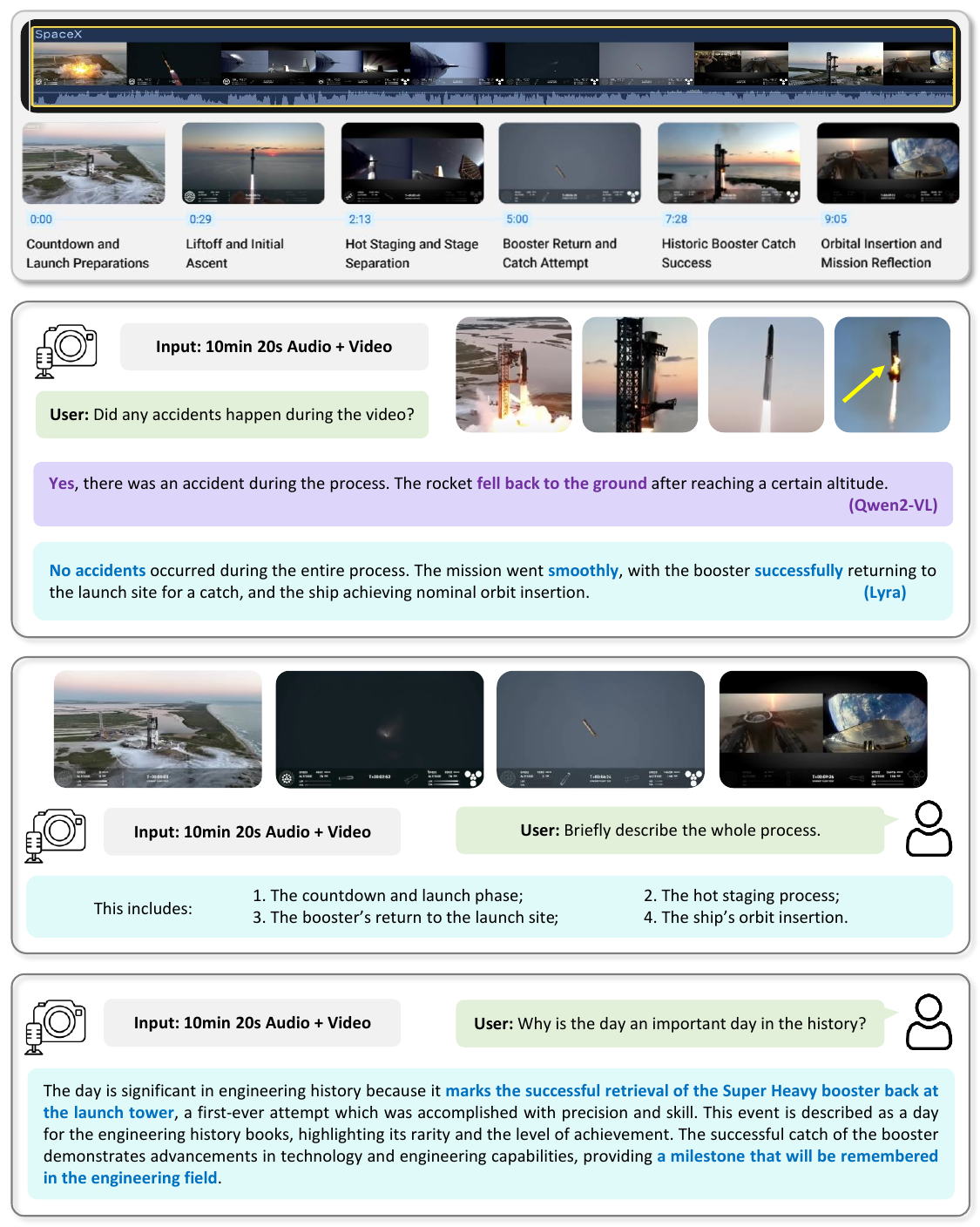}
		\vspace{-6pt}
		\caption{More long speech examples results. {\ours} achieves more accurate omni-cognition compared to naive VLMs like Qwen2-VL.}
		\label{fig:long_speech2}
		\vspace{-12pt}
	\end{center}
\end{figure*}

\begin{figure*}[!t]
	\begin{center}
		\includegraphics[width=0.9\textwidth]{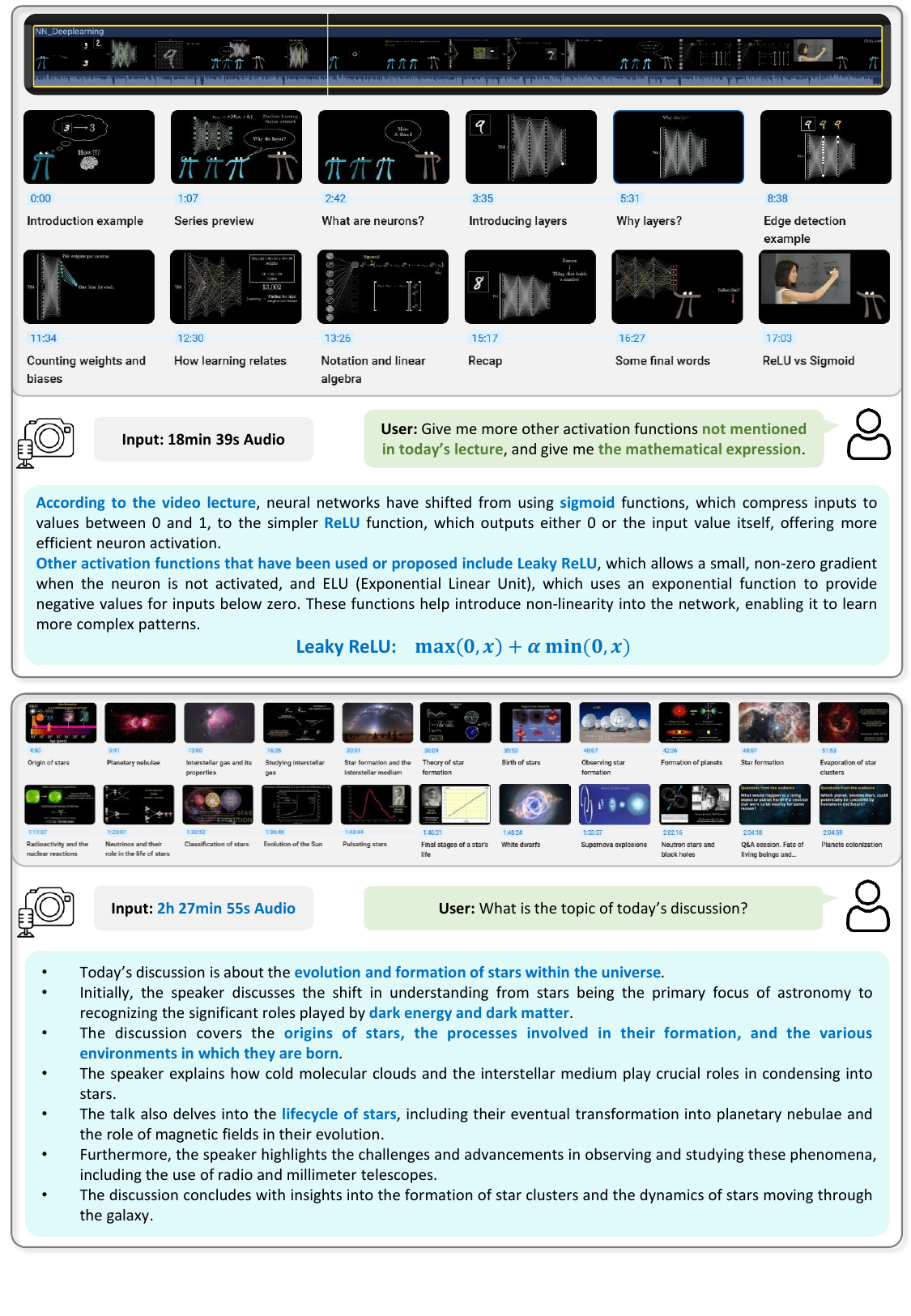}
		\vspace{-5pt}
		\caption{More examples of {\ours} with hour-long lectures (more than two hours).}
		\label{fig:long_speech3}
	\end{center}
\end{figure*}

\begin{figure*}[!t]
	\begin{center}
		\includegraphics[width=0.9\textwidth]{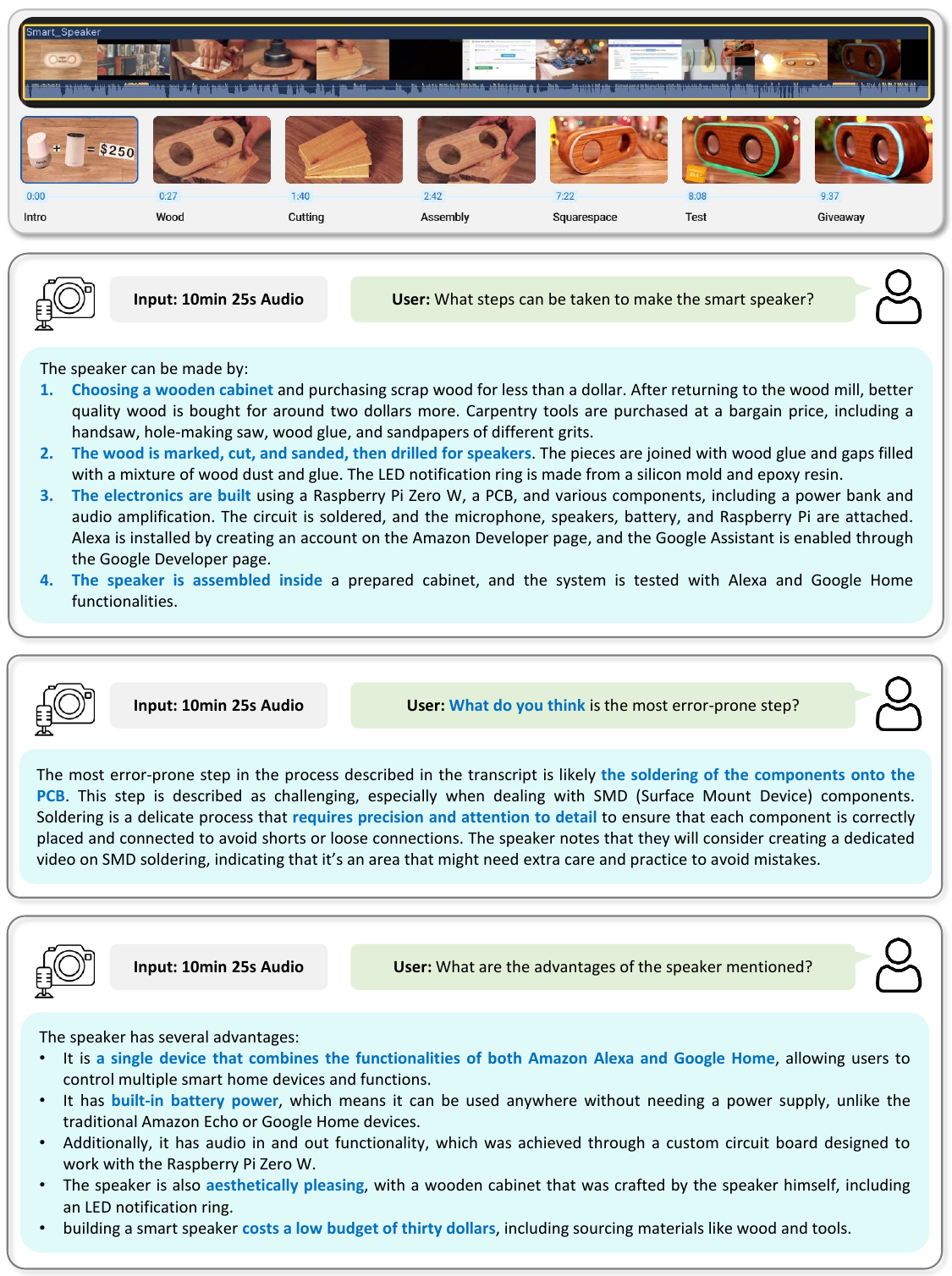}
		\vspace{-5pt}
		\caption{More results from long speech examples: Lyra can subjectively answer questions about complex steps.}
		\label{fig:long_speech4}
	\end{center}
\end{figure*}

In the main paper experimental section, Fig.~\ref{fig:needle_base} shows that existing Speech Language Models (SLMs) fail when processing audio longer than 450 seconds (about seven minutes): the output becomes nonsensical with extensive repetition. In this part, we demonstrate Lyra's ability to handle long audio inputs. In Fig.~\ref{fig:long_speech1}, \ref{fig:long_speech2}, \ref{fig:long_speech3}, and~\ref{fig:long_speech4} we demonstrate {\ours}'s capability to process long-form speech ({\color{myblue}\textbf{best view the following part together with the video in the supplementary materials}}). {\ours} effectively extracts the information that users need from extended speech contents. It excels at capturing both the details and the overall structure of long speeches. In news scenarios (Fig.~\ref{fig:long_speech1}, with frequent topic switches), it accurately identifies the focused information and responds exceptionally well.

For more complex tasks, as shown in Fig.~\ref{fig:long_speech2}, such as scenarios with visual ambiguity, our model leverages long-form speech and keyframes from videos to provide more accurate results compared to a powerful VLM like Qwen2-VL that rely solely on visual information.

In Fig.~\ref{fig:long_speech3}, our model demonstrates its ability to process daily lectures, offering significant advantages for educational-related applications. {\ours} can handle speech content durations exceeding two hours, which enables intelligent models to tackle more complex multi-modal tasks.

In Fig.~\ref{fig:long_speech4}, For tasks with longer temporal sequences and higher complexity, Lyra can also understand them and provide subjectively reasonable answers to the questions.

\clearpage
\clearpage

\clearpage

\clearpage

\clearpage

\clearpage
\twocolumn
{
	\small
	\bibliographystyle{ieeenat_fullname}
	\bibliography{main}
}


\end{document}